\pdfoutput=1

\documentclass[11pt]{article}

\usepackage[acceptedWithA]{tacl2021v1}

\usepackage[subrefformat=parens,labelformat=parens]{subfig}
\usepackage{ifthen}
\usepackage{lineno}
\usepackage{times}
\usepackage{latexsym}
\usepackage{amsfonts,amsmath,amsthm}
\usepackage{tikz}
\usepackage{mathtools}
\usepackage{multirow}
\usepackage{hhline}
\usepackage{xcolor}
\usepackage{amsmath}
\usepackage{array}

\newcolumntype{H}{>{\setbox0=\hbox\bgroup}c<{\egroup}@{}}

\usepackage{booktabs}

\usepackage{relsize}

\mathtoolsset{showonlyrefs,showmanualtags}

\definecolor{mygreenua}{HTML}{F1F5EB}
\definecolor{myredda}{HTML}{FFE6E6}

\newcommand{\uahelper}[1]{\colorbox{myredda}{\smaller$\uparrow$#1}}
\newcommand{\dahelper}[1]{\colorbox{mygreenua}{\smaller$\downarrow$#1}}

\newcommand{\uaghelper}[1]{\colorbox{mygreenua}{\smaller$\uparrow$#1}}
\newcommand{\dabhelper}[1]{\colorbox{myredda}{\smaller$\downarrow$#1}}

\newcommand{\ua}[1]{\ifthenelse{\equal{#1}{0.00} \or \equal{#1}{0.000}}{}{\uahelper{#1}}}
\newcommand{\da}[1]{\ifthenelse{\equal{#1}{0.00} \or \equal{#1}{0.000}}{}{\dahelper{#1}}}
\newcommand{\uag}[1]{\ifthenelse{\equal{#1}{0.00} \or \equal{#1}{0.000}}{}{\uaghelper{#1}}}
\newcommand{\dab}[1]{\ifthenelse{\equal{#1}{0.00} \or \equal{#1}{0.000}}{}{\dabhelper{#1}}}

\ifthenelse{\equal{1}{0}}{
 \newcommand{\onlywithappendix}[1]{}
 \newcommand{\onlywithoutappendix}[1]{#1}
}
{
\newcommand{\onlywithappendix}[1]{#1}
\newcommand{\onlywithoutappendix}[1]{}
}

\newcommand{\ignore}[1]{}

\newtheorem{theorem}{Theorem}
\newtheorem{lemma}{Lemma}
\newtheorem{proposition}{Proposition}

\usepackage{bm}

\usetikzlibrary{positioning,chains,fit,shapes,calc}
\DeclareMathOperator{\argmax}{argmax}

\usepackage[T1]{fontenc}

\usepackage[utf8]{inputenc}

\usepackage{microtype}

\newcommand{\rv}[1]{\mathbf{#1}}
\newcommand{\myvec}[1]{\mathbf{#1}}
\newcommand{\mat}[1]{\bm{#1}}
\newcommand{\mymat}[1]{\bm{#1}}
\newcommand{\elementrv}[1]{\textnormal{#1}}

\newcommand{\shaycomment}[1]{\textcolor{blue}{#1}}

\title{Erasure of Unaligned Attributes from Neural Representations}

\author{Shun Shao$^{\ast}$ $\qquad$ Yftah Ziser\Thanks{Equal contribution.} $\qquad$ Shay B. Cohen \\
Institute for Language, Cognition and Computation \\
School of Informatics, University of Edinburgh \\
10 Crichton Street, Edinburgh, EH8 9AB \\
\texttt{s.shao-11@sms.ed.ac.uk} \quad
\texttt{yftah.ziser@ed.ac.uk} \\
\texttt{scohen@inf.ed.ac.uk}}

\definecolor{mygreen}{RGB}{160,80,80}
\definecolor{myblue}{RGB}{80,160,80}
\definecolor{mybluee}{RGB}{80,80,160}

\begin{document}

\maketitle

\begin{abstract}
We present the Assignment-Maximization Spectral Attribute removaL (AMSAL) algorithm, which erases information from neural representations when the information to be erased is implicit rather than directly being aligned to each input example. Our algorithm works by alternating between two steps. In one, it finds an assignment of the input representations to the information to be erased, and in the other, it creates projections of both the input representations and the information to be erased into a joint latent space. We test our algorithm on an extensive array of datasets, including a Twitter dataset with multiple guarded attributes, the BiasBios dataset and the BiasBench benchmark. The last benchmark includes four datasets with various types of protected attributes. Our results demonstrate that bias can often be removed in our setup. We also discuss the limitations of our approach when there is a strong entanglement between the main task and the information to be erased.\footnote{Our code is available at \url{https://github.com/jasonshaoshun/AMSAL}.}
\end{abstract}

\section{Introduction}



Developing a methodology for adjusting neural representations to preserve user privacy and avoid encoding bias in them has been an active area of research in recent years. Previous work shows it is possible to erase undesired information from representations so that downstream classifiers cannot use that information in their decision-making process. This previous work assumes that this sensitive information (or \textbf{guarded} attributes, such as gender or race) is available for each input instance. These guarded attributes, however, are sensitive, and obtaining them on a large scale is often challenging and, in some cases, not feasible \cite{han2021decoupling}. For example, \newcite{blodgett2016demographic} studied the characteristics of African-American English (AAE) on Twitter, and could not couple the ethnicity attribute directly with the tweets they collected due to the attribute's sensitivity.

This paper introduces a novel debiasing setting in which the guarded attributes are not paired up with each input instance and an algorithm to remove information from representations in that setting. In our setting, we assume that each neural input representation is coupled with a guarded attribute value, but this assignment is unavailable. In cases where the domain of the guarded attribute is small (for example, with binary attributes), this means that the guarded attribute information consists of \textbf{priors} with respect to the whole population and not instance-level information.

\begin{figure}
\centering

\scalebox{0.9}{
\begin{tikzpicture}[thick,
  every node/.style={draw,circle},
  fsnode/.style={fill=myblue},
  ssnode/.style={fill=mygreen},
  every fit/.style={ellipse,draw,inner sep=-2pt,text width=2.25cm},
  ->,shorten >= 3pt,shorten <= 3pt
]

\begin{scope}[start chain=going below,node distance=7mm]
\foreach \i in {1,2,...,5}
  \node[fsnode,on chain] (f\i) [label=left: $\myvec{x}^{(\i)}$] {};
\end{scope}

\begin{scope}[xshift=5cm,yshift=-0.5cm,start chain=going below,node distance=7mm]
\foreach \i in {6,7,...,9}
\pgfmathsetmacro{\mysubtractresult}{int(\i-5)}
  \node[ssnode,on chain] (s\i) [label=right: $\myvec{z}^{(\mysubtractresult)}$] {};
\end{scope}

\node [myblue,fit=(f1) (f5),label=above:$\mymat{U}$] {};
\node [mygreen,fit=(s6) (s9),label=above:$\mymat{V}$] {};

\draw[->,line width=1.8pt,mybluee] (f1) -- (s6);
\draw (f1) -- (s7);
\draw (f1) -- (s8);
\draw (f1) -- (s9);

\draw[->,line width=1.8pt,mybluee] (f2) -- (s6);
\draw (f2) -- (s7);
\draw (f2) -- (s8);
\draw (f2) -- (s9);

\draw (f3) -- (s6);
\draw (f3) -- (s7);
\draw[->,line width=1.8pt,mybluee] (f3) -- (s8);
\draw (f3) -- (s9);

\draw (f4) -- (s6);
\draw (f4) -- (s7);
\draw (f4) -- (s8);
\draw[->,line width=1.8pt,mybluee] (f4) -- (s9);

\draw (f5) -- (s6);
\draw (f5) -- (s7);
\draw (f5) -- (s8);
\draw[->,line width=1.8pt,mybluee] (f5) -- (s9);

\end{tikzpicture}
}

\caption{\label{fig:demo} A depiction of the problem setting and solution. The inputs are aligned to each guarded sample, based on
strength using two projections $\mat{U}$ and $\mat{V}$. We solve a bipartite matching problem to find the blue edges, and then recalculate
$\mat{U}$ and $\mat{V}$.}
\end{figure}
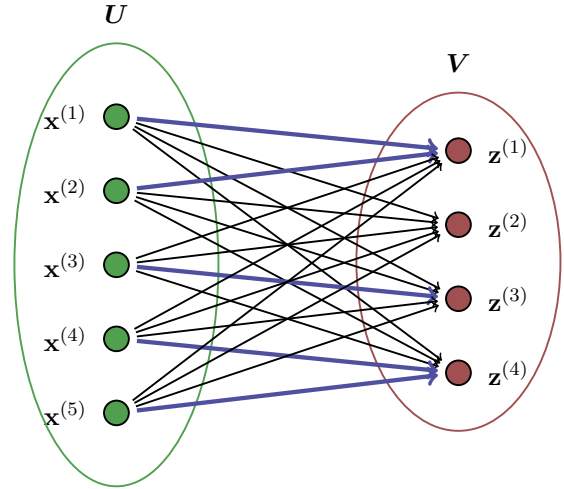

The intuition behind our algorithm is that if we were to find a strong correlation between the input variable and a set of guarded grounded attributes either in the form of an unordered list of \textbf{records} or as priors, then it is unlikely to be coincidental if the sample size is sufficiently large (\S\ref{section:justification}). We implement this intuition by jointly finding projections of the input samples and the guarded attributes into a joint embedding space \emph{and} an alignment between the two sets in that joint space.

Our resulting algorithm (\S\ref{section:algorithm}), the Alignment-Maximization Spectral Attribute removaL algorithm (AMSAL), is a coordinate-ascent algorithm reminiscent of the hard expectation-maximization algorithm (hard EM; \citealt{mackay2003information}). It first loops between two Alignment and Maximization steps, during which it finds an alignment (A) based on existing projections and then projects the representations and guarded attributes into a joint space based on an existing alignment (M). After these two steps are iteratively repeated and an alignment is identified, the algorithm takes another step to erase information from the input representations based on the projections identified. This step closely follows the work of \newcite{shao2022gold}, who use Singular Value Decomposition (SVD) to remove principal directions of the covariance matrix between the input examples and the guarded attributes. Figure~\ref{fig:demo} depicts a sketch of our setting and the corresponding algorithm, with $\rv{x}_i$ being the input representations and $\rv{z}_j$ being the guarded attributes.
 Our algorithm is modular: while our use of the algorithm of \newcite{shao2022gold} for the removal step is natural due to the nature of the AM steps, a user can use any such algorithm to erase the information from the input representations (\S\ref{section:removal-algorithm}).

Our contributions are as follows: (1) We propose a new setup for removing guarded information from neural representations where there are few or no labeled guarded attributes; (2) We present a novel two-stage coordinate-ascent algorithm that iteratively improves (a) an alignment between guarded attributes and neural representations; and (b) information removal projections.
 
 Using an array of datasets, we perform extensive experiments to assess how challenging our setup is and whether our algorithm is able to remove information without having aligned guarded attributes (\S\ref{section:experiments}). We find in several cases that little information is needed to align between neural representations and their corresponding guarded attributes. The consequence is that it is possible to erase the information such guarded attributes provide from the neural representations while preserving the information needed for the main task decision-making. We also study the limitations of our algorithm by experimenting with a setup where it is hard to distinguish between the guarded attributes and the downstream task labels when aligning the neural representations with the guarded attributes (\S\ref{section:limitations}).

\section{Problem Formulation and Notation}


For an integer $n$ we denote by $[n]$ the set $\{1,\ldots,n\}$. For a vector $\myvec{v}$, we denote by $||\myvec{v}||_2$ its $\ell_2$ norm. For two vectors $\myvec{v}$ and $\myvec{u}$, by default in column form, $\langle \myvec{v}, \myvec{u} \rangle = \myvec{v}^{\top} \myvec{u}$ (dot product). Matrices and vectors are in boldface font (with uppercase or lowercase letters, respectively). Random variable vectors are also denoted by boldface uppercase letters. For a matrix $\mat{A}$, we denote by $a_{ij}$ the value of cell
$(i,j)$. The Frobenius norm of a matrix $\mat{A}$ is $||\mat{A}||_F = \sqrt{\sum_{i,j} a_{ij}^2}$.
The spectral norm of a matrix is $||\mymat{A}||_2 = \max_{||\myvec{x}||_2=1} ||\mymat{A} \myvec{x}||_2$.
The expectation of a random variable $\rv{T}$ is denoted by $\mathbb{E}[\rv{T}]$.


In our problem formulation, we assume three random variables: $\rv{X} \in \mathbb{R}^d$, $\rv{Y} \in \mathbb{R}$ and $\rv{Z} \in \mathbb{R}^{d'}$ such that $d' \le d$ and the expectation of all three variables is $0$ (see \citealt{shao2022gold}).
Samples of $\rv{X}$ are the inputs for a classifier to predict corresponding samples of $\rv{Y}$. The random vector $\rv{Z}$ represents the guarded attributes. We want to maintain the ability to predict $\rv{Y}$ from $\rv{X}$, while minimizing the ability to predict $\rv{Z}$ from $\rv{X}$.

We assume $n$ samples of $(\rv{X}, \rv{Y})$ and $m$ samples of $\rv{Z}$, denoted by $(\myvec{x}^{(i)}, \myvec{y}^{(i)})$ for $i \in [n]$, and $\myvec{z}^{(i)}$ for $i \in [m]$ ($m \le n$). While originally, these samples were generated jointly from the underlying distribution $p(\rv{X}, \rv{Y}, \rv{Z})$,
we assume a shuffling of the $\rv{Z}$ samples in such a way that we are only left with $m$ samples that are unique (no repetitions) and an underlying unknown many-to-one
mapping $\pi \colon [n] \rightarrow [m]$ that maps each $\myvec{x}^{(i)}$
to its original $\myvec{z}^{(j)}$.

The problem formulation is such that we need to remove the information from the $x$s in such a way that we consider the samples of $z$s as a set.
In our case, we do so by iterating between trying to infer $\pi$, and then
using standard techniques, remove the information from $x$s based on their alignment to the corresponding $z$s.


\paragraph{Singular Value Decomposition} 
Let $\mat{A} = \mathbb{E}[\rv{X}\rv{Z}^{\top}]$, the matrix of cross-covariance between $\rv{X}$ and $\rv{Z}$. This means that $\mat{A}_{ij} = \mathrm{Cov}(\elementrv{X}_i,\elementrv{Z}_j)$ for $i \in [d]$ and $j \in [d']$.

For any two vectors, $\myvec{a} \in \mathbb{R}^d, \myvec{b} \in \mathbb{R}^{d'}$, the following holds due to the linearity of expectation:
\begin{align}
\myvec{a} \mat{A} \myvec{b}^{\top} = \mathrm{Cov}(\myvec{a}^{\top} \rv{X}, \myvec{b}^{\top}\rv{Z}). \label{eq:A}
\end{align}

Singular value decomposition on $\mat{A}$, in this case, finds the ``principal directions'': directions in which the projection of $\rv{X}$ and $\rv{Z}$ maximize their covariance. The projections are represented as two matrices $\mat{U} \in \mathbb{R}^{d \times d}$ and $\mat{V} \in \mathbb{R}^{d' \times d'}$. Each column in these matrices plays the role of the vectors $\myvec{a}$ and $\myvec{b}$ in Eq.~\refeq{eq:A}. SVD finds $\mat{U}$ and $\mat{V}$ such that for any $i \in [d']$ it holds that:
\begin{align}
\mathrm{Cov}(\mat{U}_i^{\top} \rv{X}, \mat{V}_i^{\top} \rv{Z})  = \max_{(\myvec{a},\myvec{b}) \in \mathcal{O}_i} \mathrm{Cov}(\myvec{a}^{\top} \rv{X}, \myvec{b}^{\top}\rv{Z}) ,
\end{align}
\noindent where $\mathcal{O}_i$ is the set of pairs of vectors $(\myvec{a},\myvec{b})$ such that $||\myvec{a}||_2 = ||\myvec{b}||_2 = 1$, $\myvec{a}$ is orthogonal to $\mat{U}_1, \ldots, \mat{U}_{i-1}$ and similarly, $\myvec{b}$ is orthogonal to $\mat{V}_1, \ldots, \mat{V}_{i-1}$. 

\newcite{shao2022gold} showed that SVD in this form can be used to debias representations. We calculate SVD between $\rv{X}$ and $\rv{Z}$ and then prune out the principal directions that denote the highest covariance.
We will use their method, SAL (Spectral Attribute removaL), in the rest of the paper. See also \S\ref{section:removal-algorithm}.

\begin{figure}[t]
\framebox{\parbox{\columnwidth}{

{\bf Inputs:} Samples $\myvec{x}^{(1)},\ldots,\myvec{x}^{(n)}$, and $\myvec{z}^{(1)},\ldots,\myvec{z}^{(m)}$.

{\bf Algorithm:} (calculate projection which removes information from the $x$s that is in the $z$s)

\medskip

Initialize $\pi$ randomly to a function from $[n]$ to $[m]$.

\medskip

Repeat the following for $T$ iterations:

\begin{itemize}

\item (M-step) Using $\pi$, let $\mat{\Omega}_{\pi}$ as in Eq.~\refeq{eq:omega}.

Calculate SVD on $\mat{\Omega}_{\pi}$ to calculate $(\mat{U}, \mat{\Sigma}, \mat{V})$.

\item (A-step) With $\mat{U}$ and $\mat{V}$ as above, with top $k$ singular vectors, find $\pi$ by solving the problem as in Eq.~\refeq{eq:opt-a-step}.

\end{itemize}

{\bf Return:} The singular vectors from $\mat{U}$ that have lowest singular values.
}}
\caption{The main Assignment-Maximization Spectral Attribute removaL (AMSAL) algorithm for removal of information without alignment between samples of $\rv{X}$ and $\rv{Z}$.}
\label{fig:ar-alg}
\end{figure}

\section{Methodology}
\label{section:algorithm}

We view the problem of information removal with unaligned samples as
a joint optimization problem of: (a) finding the alignment; (b) finding the
projection that maximizes the covariance between the alignments, and
using its complement to project the inputs.
Such an optimization, in principle, is intractable, so we break it down into
two coordinate-ascent style steps: A-step (in which the \underline{a}lignment is identified as a bipartite graph matching problem) and M-step (in which based on the previously identified alignment, a \underline{m}aximal-covariance projection is calculated). Formally, the maximization problem we solve is:

\begin{equation}
(\mat{U}, \mat{V}, \pi) = \arg\max_{\mat{U}, \mat{V}, \pi} \sum_{i=1}^n  (\myvec{x}^{(i)})^{\top} \mat{U} \mat{V}^{\top} \myvec{z}^{(i)},
\end{equation}

\noindent where we constrain $\mat{U}$ and $\mat{V}$ to be matrices
with orthonormal columns in  $\mathbb{R}^{n \times k}$.

Note that the sum in the above equation has a term per pair of $(\myvec{x}^{(i)}, \myvec{z}^{\pi(i)})$, which enables us to frame the A-step as an integer linear programming (ILP) problem (\S\ref{section:astep}).
The full algorithm is given in Figure~\ref{fig:ar-alg}, and we proceed
in the next two steps to further explain the A-step and the M-step.

\subsection{A-step (Guarded Sample Assignment)}
\label{section:astep}
\label{section:ilp}

In the Assignment Step, we are required to find a many-to-one alignment
$\pi \colon [n] \rightarrow [m]$ between $\{\myvec{x}^{(1)} , \ldots, \myvec{x}^{(n)} \}$ and $\{ \myvec{z}^{(1)}, \ldots, \myvec{z}^{(m)} \}$. Given $\mat{U}$ and $\mat{V}$ from the previous M-step, we can find such an assignment by solving the following optimization problem:

\begin{equation}
    \arg\max_{\pi} \sum_{i=1}^n \langle \mat{U}^{\top} \myvec{x}^{(i)}, \mat{V}^{\top} \myvec{z}^{(\pi(i))} \rangle. 
\end{equation}

This maximization problem can be formulated as an integer linear program of the following form:

\begin{align}
    \max_{\mymat{P} \in \{ 0, 1 \}^{n \times m} } & & \sum_{j=1}^m \sum_{i=1}^n p_{ij} \langle \mat{U}^{\top} \myvec{x}^{(i)}, \mat{V}^{\top} \myvec{z}^{(j)} \rangle \\
    \textit{s.t. } & \forall i. & \sum_{j=1}^m p_{ij} = 1, \\
    & \forall j. & b_{0j} \le \sum_{i=1}^m p_{ij} \le b_{1j}.
\label{eq:opt-a-step}
\end{align}

This is a solution to an assignment problem \cite{Kuhn1955,ramshaw2012minimum}, where $p_{ij}$ denotes whether $\myvec{x}^{(i)}$ is associated with the (type of) guarded
attribute $\myvec{z}^{(j)}$. The values $(b_{0j},b_{1j})$ determine lower
and upper bounds on the number of $x$s a given $\myvec{z}^{(j)}$ can be assigned to. While a standard assignment problem can be solved efficiently using the Hungarian method of \newcite{Kuhn1955}, we choose to use the ILP formulation, as it enables us to have more freedom in adding constraints to the problem, such as the lower and upper bounds. 

\subsection{M-step (Covariance Maximization)}

The result of an A-step is an assignment $\pi$ such that
$\pi(i) = j$ implies $\myvec{x}^{(i)}$ was deemed as aligned
to $\myvec{z}_j$. With that $\pi$ in mind, we define
the following empirical covariance matrix $\mat{\Omega}_{\pi} \in \mathbb{R}^{d \times d'}$:

\begin{equation}
    \mat{\Omega}_{\pi} = \sum_{i=1}^n \myvec{x}^{(i)} (\myvec{z}^{(\pi(i))})^{\top}.
\label{eq:omega}
\end{equation}

We then apply SVD on $\mat{\Omega}_{\pi}$ to get new $\mat{U}$ and $\mat{V}$ that are used in the next iteration of the algorithm with the A-step, if the algorithm continues to run. When the maximal number of iterations is reached, we follow the work of \newcite{shao2022gold} in using a truncated part of $\mat{U}$ to remove the information from the $x$s.  We do that by projecting $\myvec{x}^{(i)}$ using the singular vectors of $\mat{U}$ with the smallest singular values. These projected vectors co-vary the least with the guarded attributes, assuming the assignment in the last A-step was precise. This method has been shown by \newcite{shao2022gold} to be highly effective and efficient in debiasing neural representations.

\subsection{A Matrix Formulation of the AM Steps}

Let $\myvec{e}_1, \ldots, \myvec{e}_m$ be the standard basis vectors.
This means $\myvec{e}_i$ is a vector of length $m$ with $0$
in all coordinates except for the $i$th coordinate, where it is $1$.

Let $\mathcal{E}$ be the set of all matrices $\mymat{E}$ where each $\mymat{E} \in \mathcal{E}$ is such that $\mymat{E} \in \mathbb{R}^{n \times m}$ and each row is one of $\myvec{e}_i$, $i \in [m]$. In that case, $\mymat{E}\mymat{Z}^{\top}$ is an $n \times d'$ matrix, such that the $j$th row is a copy of the $i$th column of $\mymat{Z} \in \mathbb{R}^{d' \times n}$.  Therefore, the AM steps can be viewed
as solving the following maximization problem using coordinate ascent:

\begin{equation}
    \underset{\mymat{E} \in \mathcal{E}, \mymat{U}, \mymat{V}, \mymat{\Sigma}}{\argmax} || \mymat{U}^{\top} \mymat{\Sigma} \mymat{V} - \mymat{X} \mymat{E} \mymat{Z}^{\top}||_F^2,
\end{equation}

\noindent where $\mymat{U}$, $\mymat{V}$ are orthonormal matrices, and $\mymat{\Sigma}$ is a diagonal matrix with non-negative elements. This corresponds to the SVD of the matrix $\mymat{X} \mymat{E} \mymat{Z}^{\top}$.

In that case, the matrix $\mymat{E}$ can be directly mapped to an assignment in the form of $\pi$, where $\pi(i)$ would be the $j$ such that the $j$th coordinate in the $i$th row of $\mymat{E}$ is non-zero.


\subsection{Removal Algorithm}
\label{section:removal-algorithm}

The AM steps are best suited for the removal of information through SVD with an algorithm such as SAL. This is because the AM steps are optimizing an objective of the same type of SAL -- relying on the projections $\mat{U}$ and $\mat{V}$ to project the inputs and guarded representations into a joint space. However, a by-product of the algorithm in Figure~\ref{fig:ar-alg} is an assignment function $\pi$ that aligns between the inputs and the guarded representations.

With that assignment, other removal algorithms can be used, for example, the algorithm of \newcite{ravfogel2020null}. We experiment with this idea in \S\ref{section:experiments}.

\subsection{Justification of the AM Steps}
\label{section:justification}

Next, we justify our algorithm (which may be skipped on the first reading). Our justification is based on the observation that if indeed
$\rv{X}$ and $\rv{Z}$ are linked together (this connection is formalized as a latent variable in their joint distribution), then for a given sample that is permuted, the singular values of $\mymat{\Omega}$ will be larger the closer the permutation is to the identity permutation. This justifies finding such a permutation that maximizes the singular values in an SVD of $\mymat{\Omega}$.

\paragraph{More Details}
Let $\iota \colon [n] \rightarrow [n]$ be the identity permutation, $\iota(i)=i$. We will assume the case in which $n=m$ (but the justification can be generalized to the case $m < n$), and that the underlying joint distribution $p(\rv{X},\rv{Z})$ is mediated by a latent variable $\rv{H}$, such that

\begin{equation}
p(\rv{X}, \rv{Z}, \rv{H}) = p(\rv{H}) p(\rv{X} \mid \rv{H}) p(\rv{Z} \mid \rv{H}). \label{eq:joint}
\end{equation}

This implies there is a latent variable that connects $\rv{X}$ and $\rv{Z}$, and that the joint distribution $p(\rv{X},\rv{Z})$ is a mixture through $\rv{H}$.

\begin{proposition}[informal]
\label{proposition:A}
Let $\{(\myvec{x}^{(i)}, \myvec{z}^{(i)} )\}$ be a sample of size $n$ from the distribution in Eq.~\refeq{eq:joint}. Let $\pi$ be a permutation over $[n]$ uniformly sampled from the set of permutations.
Then with high likelihood, the sum of the singular values of $\mymat{\Omega}_{\pi}$ is smaller than the sum of singular values under $\mymat{\Omega}_{\iota}$.
\end{proposition}

\onlywithappendix{For full details of this claim, see Appendix~\ref{appendix:just}.}
\onlywithoutappendix{Full details of this claim will be included in an appendix.}

\ignore{

we will refer to sigma something as the sum of the singular values of the expectation. we assume X * Z' is smaller than C.

\begin{lemma}
Let $\myvec{x}^{(i)}$ be a set of i.i.d. samples and $\myvec{z}^{(j)}$ also a set of i.i.d. samples, where there is an independence also between $\myvec{x}^{(i)}$ and $\myvec{z}^{(j)}$.

be a set of i.i.d. samples in the sense that the $x$ and the $h$ come with a different $h$. with $k$ example being dependent. Then the sum of singular values .. is smaller than ...
\end{lemma}

\begin{proof}
The expectation of the outer product $\rv{X}\rv{Z}^{\top}$ where $\rv{X}$ and $\rv{Z}$ are independent and with mean zero is a zero matrix.
For $\mymat{\Omega}_{\iota}$ defined as in Eq.~\refeq{eq:omega} with the identity permutation $\iota$, it holds that \cite{}
its singular values are such that $\sigma_i - 0 \le || \mymat{\Omega}_{\iota}$.
\end{proof}

proof: | sigma - 0 | <= k/n || X * Z' || for the k samples

\begin{lemma}
Let $x^{(i)}$ and $z^{(i)}$ be i.i.d. samples. Then the singular value is larger than the singular value of the expectation minus epsilon
\end{lemma}

\begin{theorem}
Let $\myvec{x}^{(i)}$ and $\myvec{z}^{(i)}$ be a set of examples for $i \in [n]$, and let $\pi$ be uniformly sampled.
Assuming $\pi$ has $k$ identity elements, then with a certain probability,
the sum of singular values of the dependent case is larger than the sum of the non-dependent case.
\end{theorem}

note that we have k/n || X*Z'||
}

\section{Experiments}
\label{section:experiments}


In our experiments, we test several combinations of algorithms. We use the $k$-means (\textsc{KMeans}) as a substitute for the AM steps as a baseline for the assignment step of $x$s to $z$s. In addition, for the removal step (once an assignment has been identified), we test two algorithms: \textsc{SAL} (\citealt{shao2022gold}; resulting in \textsc{AMSAL}) and \textsc{INLP} \cite{ravfogel2020null}.
We also compare these two algorithms in \emph{oracle} mode (in which the assignment of guarded attributes to inputs is known), to see the loss in performance that happens due to noisy assignments from the AM or $k$-means algorithm (\textsc{OracleSAL} and \textsc{OracleINLP}).

When running the AM algorithm or $k$-means, we execute it with three random seeds (see also \S\ref{section:stability}) for a maximum of a hundred iterations and choose the projection matrix with the largest objective value over all seeds and iterations. For the slack variables ($b_{0j}$ and $b_{1j}$ variables in Eq.~\refeq{eq:opt-a-step}), we use 20\%-30\% above and below the baseline of the guarded attribute priors according to the training set. With the SAL methods, we remove the number of directions according to the rank of the $\mymat{\Omega}$ matrix (between $2$ to $6$ in all experiments overall).

In addition, we experiment with a partially supervised assignment process, in which a small seed dataset of aligned $x$s and $z$s is provided to the AM steps. We use it for model selection: rather than choosing the assignment with the highest SVD objective value, we choose the assignment with the highest accuracy on this seed dataset. We refer to this setting as \textsc{Partial} (for ``partially supervised assignment'').

Finally, in the case of a gender-protected attribute, we compare our results against a baseline in which the input $\myvec{x}$ is compared against a list of words stereotypically associated with the genders of male or female.\footnote{\url{https://tinyurl.com/33bzddtw}} Based on the overlap with these two lists, we heuristically assign the gender label to $\myvec{x}$ and then run SAL or INLP (rather than using the AM algorithm). While this wordlist heuristic is plausible in the case of gender, it is not as easy to derive in the case of other protected attributes, such as age or race. We give the results for this baseline using the marker \textsc{WL} in the corresponding tables.

\paragraph{Main Findings}
Our overall main finding shows that our novel setting in which guarded information is erased from individually-unaligned representations is viable. We discovered that AM methods perform particularly well when dealing with more complex bias removal scenarios, such as when multiple guarded attributes are present. We also found that having similar priors for the guarded attributes and downstream task labels may lead to poor performance on the task at hand. In these cases, using a small amount of supervision often effectively helps reduce bias while maintaining the utility of the representations for the main classification of the regression problem. Finally, our analysis of alignment stability shows that our AM algorithm often converges to suitable solutions that align $\rv{X}$ with $\rv{Z}$.

Due to the unsupervised nature of our problem setting, we advise validating the utility of our method in the following way. Once we run the AM algorithm, we check whether there is a high-accuracy alignment between $\rv{X}$ and $\rv{Y}$ (rather than $\rv{Z}$, which is unavailable). If this alignment is accurate, then we run the risk of significantly damaging task performance. An example is given in \S\ref{section:limitations}.

\subsection{Word Embedding Debiasing}
\label{section:experiments-we}

As a preliminary assessment of our setup and algorithms, we apply our methods to GloVe word embeddings to remove gender bias, and following the previous experiment settings of this problem \cite{bolukbasi2016man,ravfogel2020null,shao2022gold}.  We considered only the 150,000 most common words to ensure the embedding quality and omitted the rest.  We sort the remaining embeddings by their projection on the $\overrightarrow{\text{he}}$-$\overrightarrow{\text{she}}$ direction. Then we consider the top 7,500 word embeddings as male-associated words ($z=1$) and the bottom 7,500 as female-associated words ($z=-1$).

Our findings are that both the $k$-means and the AM algorithms perfectly identify
the alignment between the word embeddings and their associated gender label (100\%).
Indeed, the dataset construction itself follows
a natural perfect  clustering that these algorithms easily discover. Since the alignments are perfectly identified, the results of predicting the gender
from the word embeddings after removal are identical to the oracle case. These results are quite
close to the results of a random guess, and we refer the reader to \newcite{shao2022gold} for details
on experiments with SAL and INLP for this dataset. Considering Figure~\ref{fig:clustering-wordembeddings}, it is evident
that our algorithm essentially follows a natural clustering of the word embeddings into two clusters, female and male, as the embeddings are highly separable in this case. This is why the alignment score of $\rv{X}$ (embedding) to $\rv{Z}$ (gender) is perfect in this case. This finding indicates that this standard word embedding dataset used for debiasing is \emph{trivial to debias} -- debiasing can be done even without knowing the identity of the stereotypical gender associated with each word.


\ignore{
\begin{figure}[t]
    \centering
    \begin{tabular}{cc}
    (a) before debiasing \\
    \includegraphics[trim={0cm 0cm 0cm 1.29cm},clip,width=1.4in]{figures/embeddings_Original.pdf} &
    (b) after debiasing \\
    \includegraphics[trim={0cm 0cm 0cm 1.29cm},clip,width=1.4in]{figures/new_figures/embeddings_Projected (k=2).pdf} \\
    (a) & (b)
    \end{tabular}
    \caption{A t-SNE visualization of the word embeddings before
    any gender information removal, and after it. In (a) that the embeddings naturally
    cluster into the corresponding gender.}
    \label{fig:clustering-wordembeddings}
\end{figure}
}

\begin{figure}[t]
    \centering

    \subfloat[before debiasing]{\includegraphics[trim={0cm 0cm 0cm 1.4cm},clip,width=0.47\textwidth]{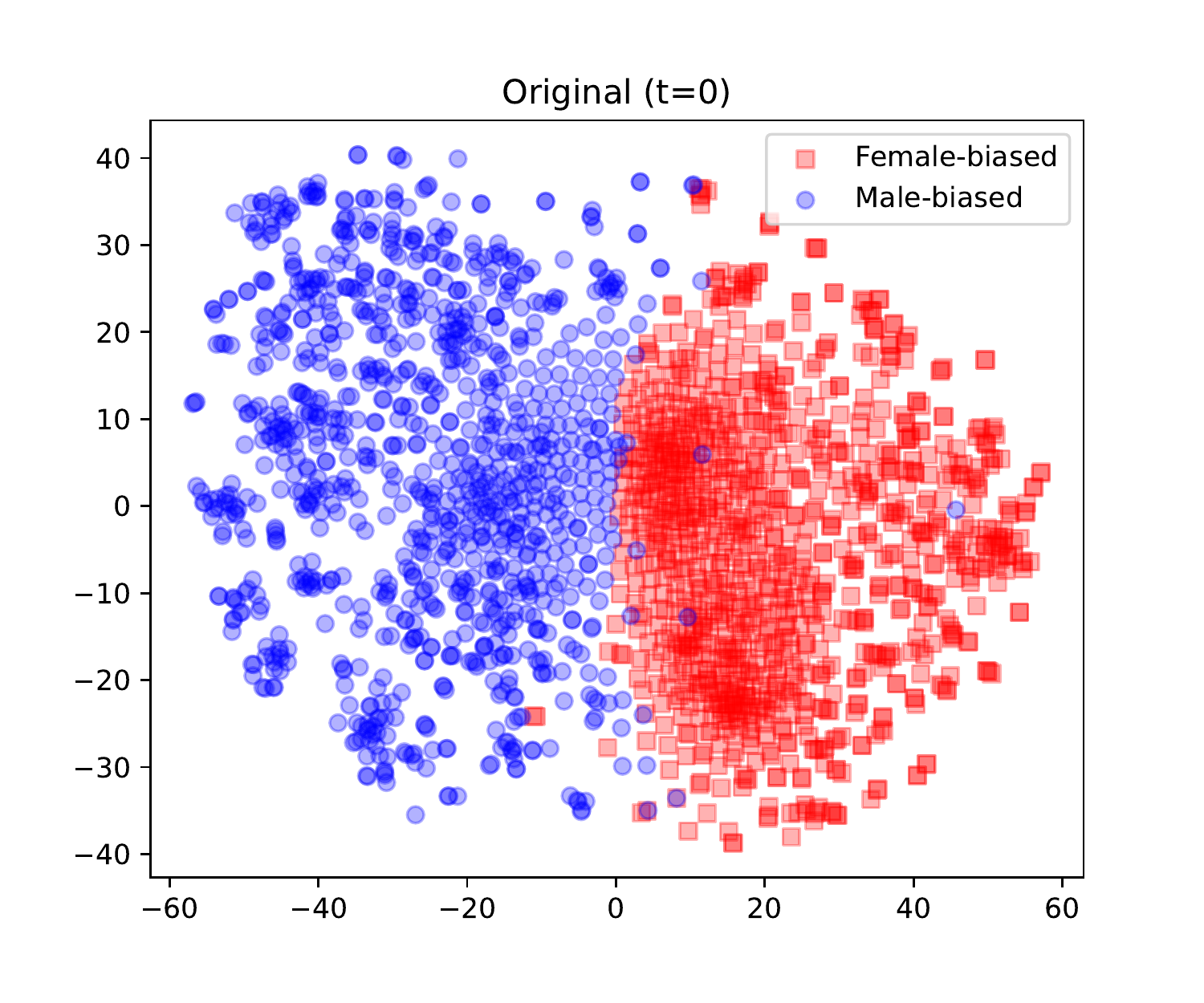}}
    \\
    \subfloat[after debiasing]{\includegraphics[trim={0cm 0cm 0cm 1.4cm},clip,width=0.47\textwidth]{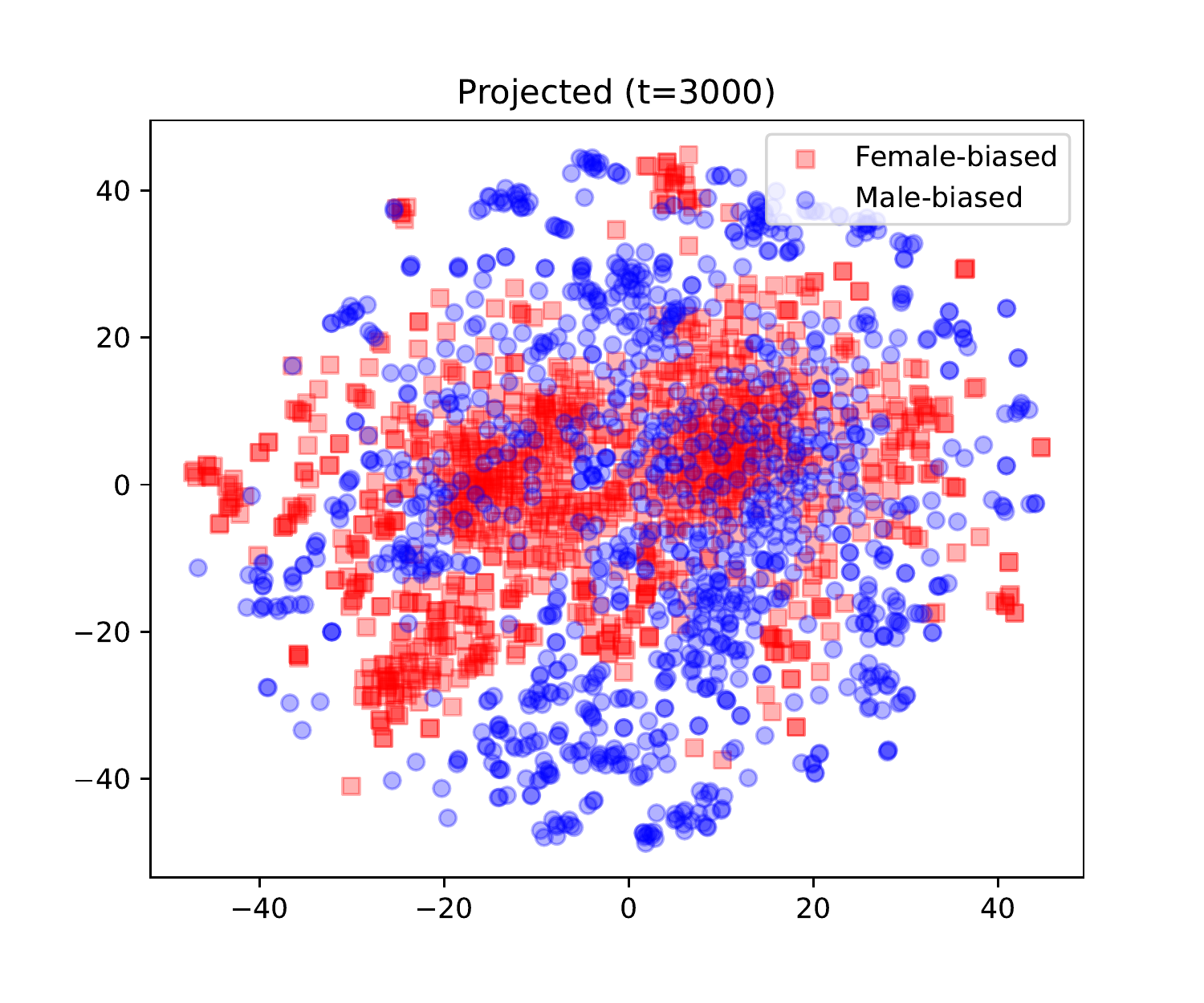}}
    \\

    \caption{A t-SNE visualization of the word embeddings before
    and after gender information removal. In (a) we see the embeddings naturally
    cluster into the corresponding gender.}
    \label{fig:clustering-wordembeddings}
\end{figure}

\begin{table}[th!]
\centering
\scalebox{0.9}{
\begin{tabular}{lrHrHH}
\toprule
                    Model &  Task Acc. &   tpr-GAP (weighted) & TPR-GAP &  f1 macro &  f1 micro \\
\midrule
                   \textsc{BertModel} &             0.79 &                 0.12 &                            0.20 &                  0.71 &                  0.79 \\
                 \, + \textsc{AMINLP} &  \dab{0.12} 0.67 &       \da{0.06} 0.06 &                  \da{0.12} 0.09 &   \dab{0.2540} 0.4599 &   \dab{0.1203} 0.6712 \\
 \, + \textsc{Kmeans} + \textsc{INLP} &  \dab{0.11} 0.68 &       \da{0.06} 0.05 &                  \da{0.12} 0.08 &   \dab{0.2328} 0.4810 &   \dab{0.1132} 0.6783 \\
 \, + \textsc{OracleINLP} &  \dab{0.11} 0.68 &       \da{0.06} 0.06 &                  \da{0.12} 0.08 &   \dab{0.2362} 0.4776 &   \dab{0.1069} 0.6846 \\
\, + \textsc{PartialINLP} &  \dab{0.12} 0.67 &       \da{0.06} 0.06 &                  \da{0.13} 0.08 &   \dab{0.2483} 0.4655 &   \dab{0.1198} 0.6718 \\
                  \, + \textsc{AMSAL} &  \uag{0.00} 0.79 &       \da{0.02} 0.10 &                  \da{0.02} 0.18 &   \uag{0.0026} 0.7165 &   \uag{0.0025} 0.7940 \\
  \, + \textsc{Kmeans} + \textsc{SAL} &  \uag{0.00} 0.79 &       \da{0.02} 0.10 &                  \da{0.02} 0.18 &   \uag{0.0022} 0.7160 &   \uag{0.0022} 0.7937 \\
  \, + \textsc{OracleSAL} &  \uag{0.00} 0.79 &       \da{0.02} 0.10 &                  \da{0.02} 0.18 &   \uag{0.0020} 0.7159 &   \uag{0.0021} 0.7936 \\
 \, + \textsc{PartialSAL} &  \uag{0.00} 0.79 &       \da{0.02} 0.10 &                  \da{0.02} 0.18 &   \uag{0.0027} 0.7165 &   \uag{0.0026} 0.7941 \\
      \, + \textsc{WL} + \textsc{SAL} &  \uag{0.00} 0.79 &       \da{0.02} 0.10 &                  \da{0.02} 0.18 &   \uag{0.0017} 0.7156 &   \uag{0.0022} 0.7938 \\
     \, + \textsc{WL} + \textsc{INLP} &  \dab{0.12} 0.68 &       \da{0.06} 0.05 &                  \da{0.12} 0.08 &   \dab{0.2438} 0.4700 &   \dab{0.1155} 0.6760 \\

\midrule
                    \textsc{FastText} &             0.77 &                 0.12 &                            0.20 &                  0.69 &                  0.77 \\
                 \, + \textsc{AMINLP} &  \dab{0.05} 0.73 &       \ua{0.00} 0.13 &                  \ua{0.01} 0.21 &   \dab{0.0612} 0.6293 &   \dab{0.0459} 0.7257 \\
 \, + \textsc{Kmeans} + \textsc{INLP} &  \dab{0.08} 0.69 &       \da{0.01} 0.12 &                  \da{0.00} 0.19 &   \dab{0.0889} 0.6016 &   \dab{0.0844} 0.6872 \\
 \, + \textsc{OracleINLP} &  \dab{0.03} 0.74 &       \da{0.07} 0.06 &                  \da{0.10} 0.09 &   \dab{0.0551} 0.6354 &   \dab{0.0321} 0.7395 \\
\, + \textsc{PartialINLP} &  \dab{0.04} 0.74 &       \da{0.03} 0.09 &                  \da{0.04} 0.16 &   \dab{0.0463} 0.6442 &   \dab{0.0353} 0.7363 \\
                  \, + \textsc{AMSAL} &  \dab{0.03} 0.74 &       \da{0.01} 0.11 &                  \da{0.03} 0.17 &   \dab{0.0207} 0.6698 &   \dab{0.0302} 0.7413 \\
  \, + \textsc{Kmeans} + \textsc{SAL} &  \dab{0.04} 0.73 &       \da{0.01} 0.11 &                  \da{0.02} 0.17 &   \dab{0.0238} 0.6667 &   \dab{0.0371} 0.7345 \\
  \, + \textsc{OracleSAL} &  \dab{0.01} 0.76 &       \da{0.06} 0.06 &                  \da{0.08} 0.12 &   \dab{0.0152} 0.6753 &   \dab{0.0112} 0.7603 \\
 \, + \textsc{PartialSAL} &  \dab{0.01} 0.76 &       \da{0.01} 0.11 &                  \da{0.02} 0.18 &   \dab{0.0062} 0.6843 &   \dab{0.0078} 0.7637 \\
      \, + \textsc{WL} + \textsc{SAL} &  \dab{0.01} 0.76 &       \da{0.06} 0.06 &                  \da{0.08} 0.12 &   \dab{0.0154} 0.6751 &   \dab{0.0108} 0.7608 \\
     \, + \textsc{WL} + \textsc{INLP} &  \dab{0.03} 0.74 &       \da{0.06} 0.06 &                  \da{0.10} 0.10 &   \dab{0.0564} 0.6341 &   \dab{0.0334} 0.7381 \\
\bottomrule
\end{tabular}
}

\caption{\label{table:biography}BiasBios dataset results. The top part uses BERT embeddings to encode the biographies, while the bottom part uses FastText embeddings.}
\end{table}

\subsection{BiasBios Results}
\newcite{de2019bias} presented the BiasBios dataset, which consists of self-provided biographies paired with the profession and gender of their authors. A list of pronouns and names is used to obtain the authors' gender automatically. They aim to expose the caveats of automated hiring systems by showing that even the simple task of predicting a candidate's profession can be affected by the candidate's gender, which is encoded in the biography representation.
For example, we want to avoid one being identified as ``he'' or ``she'' in their biography, affecting the likelihood of them being classified as engineers or teachers.

We follow the setup of \newcite{de2019bias}, predicting a candidate's professions ($\myvec{y}$), based on a self-provided short biography ($\myvec{x}$), aiming to remove any information about the candidate's gender ($\myvec{z}$). Due to computational constraints, we use only random 30K examples to learn the projections with both SAL and INLP (whether in the unaligned or aligned setting). For the classification problem, we use the full dataset. To get vector representations for the biographies, we use two different encoders, FastText word embeddings \cite{joulin2016fasttext}, and BERT \cite{devlin2018bert}. We stack a multi-class classifier on top of these representations, as there are 28 different professions.  We use 20\% of the training examples for the \textsc{Partial} setting. For BERT, we followed \newcite{de2019bias} in using the last CLS token state as the representation of the whole biography.
We used the BERT model \texttt{bert-base-uncased}. 

\paragraph{Evaluation Measures}
We use an extension of the True Positive Rate (TPR) gap, the root mean square (RMS) TPR gap of all classes, for evaluating bias in a multiclass setting. This metric was suggested by  \newcite{de2019bias}, who demonstrated it is significantly correlated with gender imbalances, which often lead to unfair classification.  The higher the metric value is, the bigger the gap between the two categories (for example, between male and female) for the specific main task prediction. For the profession classification, we report accuracy.

\paragraph{Results}
Table~\ref{table:biography} provides the results for the biography dataset. We see that INLP significantly reduces the TPR-GAP in all settings, but this comes at a cost: the representations are significantly less useful for the main task of predicting the profession.
When inspecting the alignments, we observe that their accuracy is quite high with BERT: 100\% with $k$-means, 85\% with the AM algorithm and 99\% with \textsc{Partial} AM. FastText results are lower, hovering around 55\% for all three methods. The high BERT assignment performance indicates that the BiasBios BERT representations are naturally separated by gender. 
We also observe that the results of  WL+SAL and WL+INLP are correspondingly identical to Oracle+SAL and Oracle+INLP. This comes as no surprise, as the gender label is derived from a similar word list, which enables the WL approach to get a nearly perfect alignment (over 96\% agreement with the gender label).



\begin{table*}[th!]
\centering

\begin{tabular}{ccccc}

\scalebox{0.8}{
\begin{minipage}{0.33\textwidth}

\begin{tabular}{lr}
\toprule
                  Model  & Stt. Score \\
\midrule
                    Gender \\
\midrule
                                BERT &               57.25 \\
    $\,$ + \textsc{AM} + \textsc{INLP} &     \ua{0.38} 57.63 \\
$\,$ + \textsc{Kmeans} + \textsc{INLP} &     \da{3.81} 53.44 \\
   $\,$ + \textsc{Oracle}\textsc{INLP} &     \da{4.58} 52.67 \\
  $\,$ + \textsc{Partial}\textsc{INLP} &     \da{4.58} 52.67 \\
        $\,$ + \textsc{AMSAL}&     \da{3.05} 54.20 \\
 $\,$ + \textsc{Kmeans} + \textsc{SAL} &     \da{2.29} 54.96 \\
    $\,$ + \textsc{Oracle}\textsc{SAL} &     \da{5.72} 51.53 \\
   $\,$ + \textsc{Partial}\textsc{SAL} &     \da{5.72} 51.53 \\

\midrule
                              ALBERT &               48.09 \\
    $\,$ + \textsc{AM} + \textsc{INLP} &     \ua{1.14} 46.95 \\
$\,$ + \textsc{Kmeans} + \textsc{INLP} &     \ua{0.38} 47.71 \\
   $\,$ + \textsc{Oracle}\textsc{INLP} &     \ua{4.58} 43.51 \\
  $\,$ + \textsc{Partial}\textsc{INLP} &     \ua{4.20} 43.89 \\
        $\,$ + \textsc{AMSAL}&     \ua{0.76} 47.33 \\
 $\,$ + \textsc{Kmeans} + \textsc{SAL} &     \da{0.38} 48.47 \\
    $\,$ + \textsc{Oracle}\textsc{SAL} &     \ua{0.76} 47.33 \\
   $\,$ + \textsc{Partial}\textsc{SAL} &     \ua{0.76} 47.33 \\

\midrule
                             RoBERTa &               60.15 \\
    $\,$ + \textsc{AM} + \textsc{INLP} &     \da{3.45} 56.70 \\
$\,$ + \textsc{Kmeans} + \textsc{INLP} &     \da{7.66} 52.49 \\
   $\,$ + \textsc{Oracle}\textsc{INLP} &     \da{4.98} 55.17 \\
  $\,$ + \textsc{Partial}\textsc{INLP} &     \da{4.98} 55.17 \\
        $\,$ + \textsc{AMSAL}&     \da{3.45} 56.70 \\
 $\,$ + \textsc{Kmeans} + \textsc{SAL} &     \da{3.83} 56.32 \\
    $\,$ + \textsc{Oracle}\textsc{SAL} &     \da{8.81} 48.66 \\
   $\,$ + \textsc{Partial}\textsc{SAL} &     \da{8.81} 48.66 \\
 
 \midrule
                               GPT-2 &               56.87 \\
    $\,$ + \textsc{AM} + \textsc{INLP} &     \da{6.11} 50.76 \\
$\,$ + \textsc{Kmeans} + \textsc{INLP} &     \da{2.67} 54.20 \\
   $\,$ + \textsc{Oracle}\textsc{INLP} &     \da{6.49} 50.38 \\
  $\,$ + \textsc{Partial}\textsc{INLP} &     \da{6.11} 50.76 \\
        $\,$ + \textsc{AMSAL}&     \ua{1.15} 58.02 \\
 $\,$ + \textsc{Kmeans} + \textsc{SAL} &     \da{3.05} 53.82 \\
    $\,$ + \textsc{Oracle}\textsc{SAL} &               56.87 \\
   $\,$ + \textsc{Partial}\textsc{SAL} &               56.87 \\
 
\bottomrule
\end{tabular}

\end{minipage}
}

& &

\scalebox{0.8}{

\begin{minipage}{0.33\textwidth}

\begin{tabular}{lr}
\toprule
                  Model  & Stt. Score \\
\midrule
                        Race \\
\midrule
                                BERT &               62.33 \\
    $\,$ + \textsc{AM} + \textsc{INLP} &     \da{1.75} 60.58 \\
$\,$ + \textsc{Kmeans} + \textsc{INLP} &     \ua{4.85} 67.18 \\
   $\,$ + \textsc{Oracle}\textsc{INLP} &     \ua{5.63} 67.96 \\
  $\,$ + \textsc{Partial}\textsc{INLP} &     \ua{5.63} 67.96 \\
        $\,$ + \textsc{AMSAL}&     \ua{0.19} 62.52 \\
 $\,$ + \textsc{Kmeans} + \textsc{SAL} &     \ua{0.19} 62.52 \\
    $\,$ + \textsc{Oracle}\textsc{SAL} &     \ua{0.78} 63.11 \\
   $\,$ + \textsc{Partial}\textsc{SAL} &     \ua{0.78} 63.11 \\

\midrule
                              ALBERT &               62.52 \\
    $\,$ + \textsc{AM} + \textsc{INLP} &     \ua{0.98} 36.50 \\
$\,$ + \textsc{Kmeans} + \textsc{INLP} &     \ua{3.50} 33.98 \\
   $\,$ + \textsc{Oracle}\textsc{INLP} &     \da{7.18} 55.34 \\
  $\,$ + \textsc{Partial}\textsc{INLP} &     \da{7.18} 55.34 \\
        $\,$ + \textsc{AMSAL}&     \da{5.82} 43.30 \\
 $\,$ + \textsc{Kmeans} + \textsc{SAL} &     \da{6.40} 43.88 \\
    $\,$ + \textsc{Oracle}\textsc{SAL} &     \da{3.69} 41.17 \\
   $\,$ + \textsc{Partial}\textsc{SAL} &     \da{3.69} 41.17 \\

\midrule
                             RoBERTa &               63.57 \\
    $\,$ + \textsc{AM} + \textsc{INLP} &     \da{9.31} 45.74 \\
$\,$ + \textsc{Kmeans} + \textsc{INLP} &     \da{6.79} 43.22 \\
   $\,$ + \textsc{Oracle}\textsc{INLP} &     \da{1.75} 61.82 \\
  $\,$ + \textsc{Partial}\textsc{INLP} &     \da{1.75} 61.82 \\
        $\,$ + \textsc{AMSAL}&     \ua{1.74} 65.31 \\
 $\,$ + \textsc{Kmeans} + \textsc{SAL} &     \ua{1.74} 65.31 \\
    $\,$ + \textsc{Oracle}\textsc{SAL} &     \ua{3.48} 67.05 \\
   $\,$ + \textsc{Partial}\textsc{SAL} &     \ua{3.48} 67.05 \\

 \midrule
                               GPT-2 &               59.69 \\
    $\,$ + \textsc{AM} + \textsc{INLP} &     \da{3.88} 55.81 \\
$\,$ + \textsc{Kmeans} + \textsc{INLP} &     \da{1.16} 58.53 \\
   $\,$ + \textsc{Oracle}\textsc{INLP} &     \ua{0.19} 59.88 \\
  $\,$ + \textsc{Partial}\textsc{INLP} &               59.69 \\
        $\,$ + \textsc{AMSAL}&     \da{4.65} 55.04 \\
 $\,$ + \textsc{Kmeans} + \textsc{SAL} &     \da{5.43} 45.74 \\
    $\,$ + \textsc{Oracle}\textsc{SAL} &     \da{4.85} 54.84 \\
   $\,$ + \textsc{Partial}\textsc{SAL} &     \da{4.85} 54.84 \\

\bottomrule
\end{tabular}

\end{minipage}
}

& &

\scalebox{0.8}{

\begin{minipage}{0.33\textwidth}

\begin{tabular}{lr}
\toprule
                  Model  & Stt. Score \\
\midrule
                        Religion \\
\midrule
                                BERT &               62.86 \\
    $\,$ + \textsc{AM} + \textsc{INLP} &     \ua{0.95} 63.81 \\
$\,$ + \textsc{Kmeans} + \textsc{INLP} &     \ua{4.76} 67.62 \\
   $\,$ + \textsc{Oracle}\textsc{INLP} &     \da{1.91} 60.95 \\
  $\,$ + \textsc{Partial}\textsc{INLP} &     \da{1.91} 60.95 \\
        $\,$ + \textsc{AMSAL}&     \ua{1.90} 64.76 \\
 $\,$ + \textsc{Kmeans} + \textsc{SAL} &     \ua{1.90} 64.76 \\
    $\,$ + \textsc{Oracle}\textsc{SAL} &     \ua{4.76} 67.62 \\
   $\,$ + \textsc{Partial}\textsc{SAL} &     \ua{4.76} 67.62 \\

\midrule
                              ALBERT &               60.00 \\
    $\,$ + \textsc{AM} + \textsc{INLP} &     \da{0.95} 59.05 \\
$\,$ + \textsc{Kmeans} + \textsc{INLP} &     \ua{9.52} 69.52 \\
   $\,$ + \textsc{Oracle}\textsc{INLP} &     \da{2.86} 57.14 \\
  $\,$ + \textsc{Partial}\textsc{INLP} &     \da{2.86} 57.14 \\
        $\,$ + \textsc{AMSAL}&    \ua{10.48} 70.48 \\
 $\,$ + \textsc{Kmeans} + \textsc{SAL} &     \ua{9.52} 69.52 \\
    $\,$ + \textsc{Oracle}\textsc{SAL} &     \ua{6.67} 66.67 \\
   $\,$ + \textsc{Partial}\textsc{SAL} &     \ua{6.67} 66.67 \\

\midrule
                             RoBERTa &               60.95 \\
    $\,$ + \textsc{AM} + \textsc{INLP} &     \ua{4.76} 65.71 \\
$\,$ + \textsc{Kmeans} + \textsc{INLP} &     \ua{2.86} 63.81 \\
   $\,$ + \textsc{Oracle}\textsc{INLP} &     \ua{1.91} 62.86 \\
  $\,$ + \textsc{Partial}\textsc{INLP} &     \ua{1.91} 62.86 \\
        $\,$ + \textsc{AMSAL}&    \ua{12.38} 73.33 \\
 $\,$ + \textsc{Kmeans} + \textsc{SAL} &    \ua{12.38} 73.33 \\
    $\,$ + \textsc{Oracle}\textsc{SAL} &    \ua{10.48} 71.43 \\
   $\,$ + \textsc{Partial}\textsc{SAL} &    \ua{10.48} 71.43 \\

 \midrule
                               GPT-2 &               61.90 \\
    $\,$ + \textsc{AM} + \textsc{INLP} &               61.90 \\
$\,$ + \textsc{Kmeans} + \textsc{INLP} &               61.90 \\
   $\,$ + \textsc{Oracle}\textsc{INLP} &               61.90 \\
  $\,$ + \textsc{Partial}\textsc{INLP} &               61.90 \\
        $\,$ + \textsc{AMSAL}&     \ua{3.81} 65.71 \\
 $\,$ + \textsc{Kmeans} + \textsc{SAL} &     \da{1.90} 60.00 \\
    $\,$ + \textsc{Oracle}\textsc{SAL} &    \ua{15.24} 77.14 \\
   $\,$ + \textsc{Partial}\textsc{SAL} &    \ua{15.24} 77.14 \\

\bottomrule
\end{tabular}

\end{minipage}
}
\\

(a) & & (b) & & (c)

\end{tabular}

\caption{(a) CrowS-Pairs Gender stereotype scores (Stt. score) in language models debiased by different debiasing techniques and assignment; (b) CrowS-Pairs Race stereotype scores; (c) CrowS-Pairs Religion stereotype scores. All models are deemed least biased if the stereotype score is 50\%. The colored numbers are calculated as $|\, | b- 50| - |s - 50| \, |$ where $b$ is the top row score and $s$ is the corresponding system score.\label{table:crows-gender}}
\end{table*}

\begin{table}[th!]
\scalebox{0.87}{
\begin{tabular}{lrr}
\toprule
                   Model & S. Score (\%) &  LM Score (\%) \\
\midrule
                                    BERT &                   60.28 &                       84.17 \\
        $\,$ + \textsc{AM} + \textsc{INLP} &         \da{1.14} 59.14 &            \dab{0.43} 83.75 \\
    $\,$ + \textsc{Kmeans} + \textsc{INLP} &         \da{0.16} 60.12 &            \dab{0.47} 83.70 \\
       $\,$ + \textsc{OracleINLP} &         \da{2.93} 57.35 &            \dab{1.07} 83.11 \\
      $\,$ + \textsc{PartialINLP} &         \da{2.93} 57.35 &            \dab{1.07} 83.10 \\
    $\,$ + \textsc{AMSAL}&         \ua{0.61} 60.89 &            \uag{0.09} 84.26 \\
$\,$ + \textsc{Kmeans} + \textsc{SAL} &         \ua{0.19} 60.47 &            \uag{0.13} 84.30 \\
   $\,$ + \textsc{OracleSAL} &         \da{0.83} 59.44 &            \uag{0.53} 84.70 \\
  $\,$ + \textsc{PartialSAL} &         \da{0.83} 59.44 &            \uag{0.53} 84.70 \\
  
\midrule
                                  ALBERT &                   59.93 &                       89.77 \\
        $\,$ + \textsc{AM} + \textsc{INLP} &         \da{0.29} 59.64 &            \dab{1.45} 88.32 \\
    $\,$ + \textsc{Kmeans} + \textsc{INLP} &         \da{0.59} 59.34 &            \dab{0.08} 89.69 \\
       $\,$ + \textsc{OracleINLP} &         \da{2.73} 57.20 &            \dab{1.59} 88.17 \\
      $\,$ + \textsc{PartialINLP} &         \da{2.72} 57.21 &            \dab{1.62} 88.15 \\
    $\,$ + \textsc{AMSAL}&         \da{0.22} 59.71 &            \dab{0.32} 89.45 \\
$\,$ + \textsc{Kmeans} + \textsc{SAL} &         \ua{0.56} 60.49 &            \dab{0.10} 89.67 \\
   $\,$ + \textsc{OracleSAL} &         \da{2.18} 57.75 &            \dab{0.16} 89.61 \\
  $\,$ + \textsc{PartialSAL} &         \da{2.18} 57.75 &            \dab{0.16} 89.61 \\
  
\midrule
                                 RoBERTa &                   66.32 &                       88.95 \\
        $\,$ + \textsc{AM} + \textsc{INLP} &         \da{4.95} 61.37 &            \uag{0.04} 88.99 \\
    $\,$ + \textsc{Kmeans} + \textsc{INLP} &         \da{2.20} 64.13 &            \dab{1.47} 87.48 \\
       $\,$ + \textsc{Oracle}\textsc{INLP} &         \da{3.82} 62.51 &            \dab{0.92} 88.03 \\
      $\,$ + \textsc{Partial}\textsc{INLP} &         \da{3.82} 62.51 &            \dab{0.91} 88.04 \\
    $\,$ + \textsc{AMSAL}&         \da{0.63} 65.70 &            \uag{0.60} 89.54 \\
$\,$ + \textsc{Kmeans} + \textsc{SAL} &         \da{0.49} 65.83 &            \uag{0.46} 89.41 \\
   $\,$ + \textsc{OracleSAL} &         \da{3.32} 63.00 &            \uag{0.40} 89.35 \\
  $\,$ + \textsc{PartialSAL} &         \da{3.32} 63.00 &            \uag{0.40} 89.35 \\
  
\midrule
                                   GPT-2 &                   62.65 &                       91.01 \\
        $\,$ + \textsc{AM} + \textsc{INLP} &         \da{1.65} 61.00 &            \dab{3.77} 87.24 \\
    $\,$ + \textsc{Kmeans} + \textsc{INLP} &         \da{1.57} 61.08 &            \dab{3.09} 87.93 \\
       $\,$ + \textsc{Oracle}\textsc{INLP} &         \da{1.26} 61.39 &            \dab{0.00} 91.01 \\
      $\,$ + \textsc{Partial}\textsc{INLP} &         \da{1.26} 61.39 &            \dab{0.00} 91.01 \\
    $\,$ + \textsc{AMSAL}&         \da{1.58} 61.07 &            \dab{0.23} 90.79 \\
$\,$ + \textsc{Kmeans} + \textsc{SAL} &         \da{4.00} 58.64 &            \dab{0.60} 90.41 \\
   $\,$ + \textsc{OracleSAL} &         \da{4.55} 58.09 &            \dab{1.75} 89.26 \\
  $\,$ + \textsc{PartialSAL} &         \da{4.55} 58.09 &            \dab{1.75} 89.26 \\

\bottomrule
\end{tabular}
}
\caption{StereoSet stereotype scores (Stt. Score) and language modeling scores (LM Score) for the gender category.  Stereotype scores indicate the least bias at 50\%, and the LM scores indicate high usability at 100\%.\label{table:stereoset-gender}}
\end{table}

\subsection{BiasBench Results}


\newcite{meade2022empirical} followed an empirical study of an array of datasets
in the context of debiasing. They analyzed different methods
and tasks, and we follow their benchmark evaluation to assess
our AMSAL algorithm and other methods in the context of our
new setting. We include a short description of the
datasets we use in this section.
\onlywithappendix{We include full results in Appendix~\ref{appendix:benchbias}, with a description
of other datasets.} We \onlywithappendix{also } 
encourage the reader to refer to \newcite{meade2022empirical}
for details on this benchmark. We use 20\% of the training examples for the \textsc{Partial} setting.




\paragraph{StereoSet \cite{nadeem-etal-2021-stereoset}}
This dataset presents a word completion test for a language model, where the completion can be stereotypical or non-stereotypical. The bias is then measured by calculating how often a model prefers the stereotypical completion over the non-stereotypical one. \newcite{nadeem-etal-2021-stereoset} introduced the language model score to measure the language model usability, which is the percentage of examples for which a model prefers the stereotypical or non-stereotypical word over some unrelated word.


\paragraph{CrowS-Pairs \cite{nangia2020crows}} This dataset includes pairs of sentences that are minimally different at the token level, but these differences lead to the sentence being either stereotypical or anti-stereotypical. The assessment measures how many times a language model prefers the stereotypical element in a pair over the anti-stereotypical element.

\paragraph{Results}
We start with an assessment of the BERT model for the CrowS-Pairs gender, race and religion bias evaluation (Table~\ref{table:crows-gender}).
We observe that all approaches for gender, except \textsc{AM+INLP} reduce the stereotype score. Race and religion are more difficult to debias in the case of BERT. INLP with $k$-means works best when no seed alignment data is provided at all, but when we consider \textsc{PartialSAL}, in which we use the alignment algorithm with some seed aligned data, we see that the results are the strongest. When we consider the RoBERTa model, the results are similar, with \textsc{PartialSAL} significantly reducing the bias. 
Our findings from Table~\ref{table:crows-gender} overall indicate that the ability to debias a representation \emph{highly depends on the model that generates the representation}. \onlywithappendix{In Table~\ref{tab:glue} we observe that the representations, on average, are not damaged for most GLUE tasks.}\onlywithoutappendix{Additional analysis, included with a full version appendix, shows that the representations, on average, \emph{are not damaged for most GLUE tasks}.}

As \newcite{meade2022empirical} have noted, when changing the representations of a language model to remove bias, we might cause such adjustments that damage the usability of the language model. To test which methods possibly cause such an issue, we also assess the language model score on the StereoSet
dataset in Table~\ref{table:stereoset-gender}. We overall see that often SAL-based methods give lower stereotype score, while INLP methods more significantly damage the language model score. This implies that the \emph{SAL-based methods remove bias effectively while less significantly harming the usability of the language model representations}.

We also conducted comprehensive results for other datasets (SEAT and GLUE) and categories of bias (based on race and religion). The results, especially for GLUE,
demonstrate the effectiveness of our method of unaligned information removal. For GLUE, we consistently retain the baseline task performance almost in full. \onlywithappendix{See Appendix~\ref{appendix:benchbias}.}\onlywithoutappendix{\footnote{The tables of results will be included in the final version.}}

\subsection{Multiple-Guarded Attribute Sentiment}
\label{section:ideology-dataset}

\begin{figure}[t]
    \centering
    \includegraphics[width=2.8in]{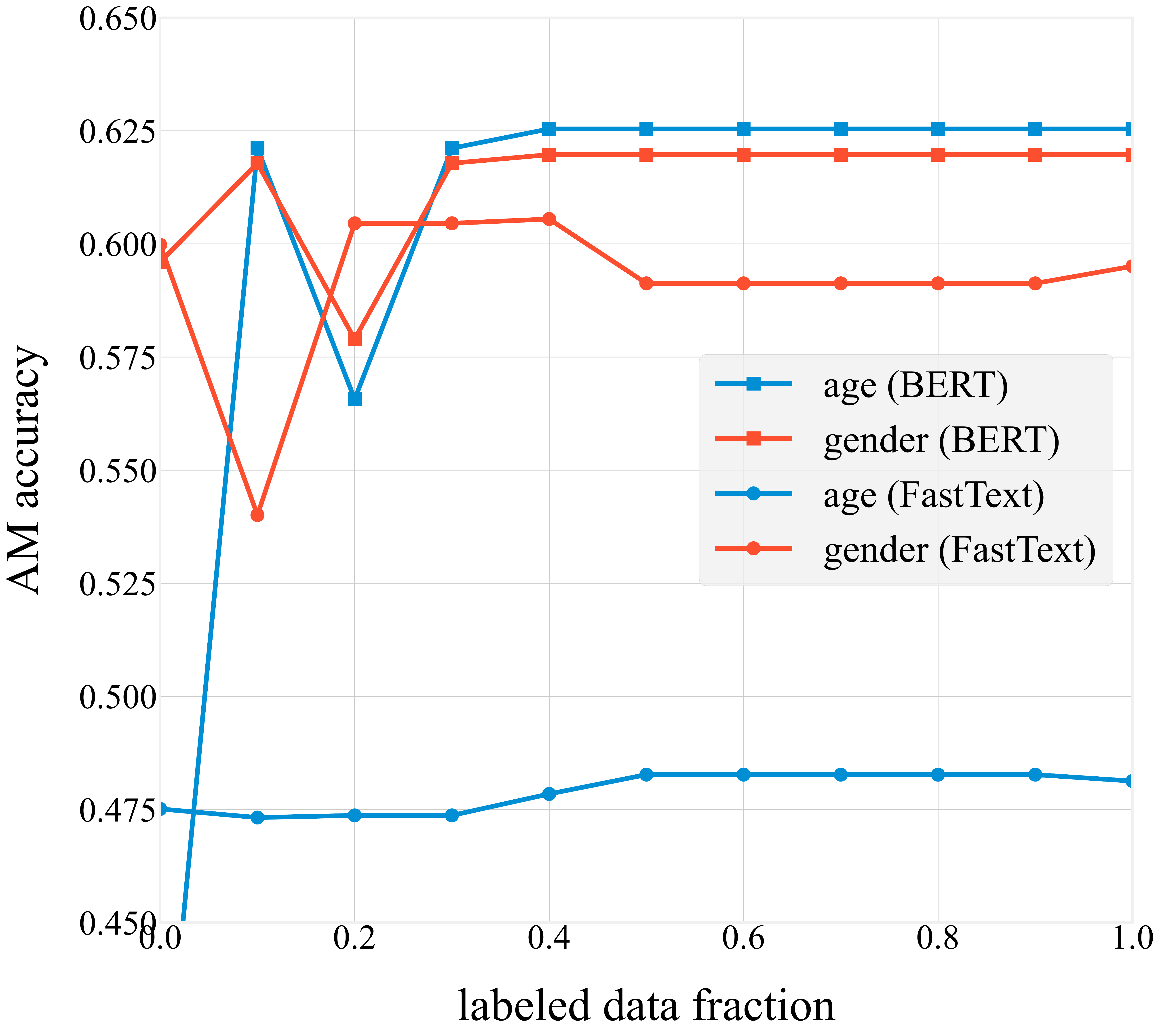}
    \caption{Accuracy of the AM steps with respect to age and gender separately (on unseen data), as a function of the fraction of the labeled dataset used by the AM algorithm.}
    \label{fig:twitter-age-gender}
\end{figure}
We hypothesize that AM-based methods are better suited for setups where multiple guarded attributes should be removed, as they allow us to target several guarded attributes with different priors. To examine our hypothesis, we experiment with a dataset curated from Twitter (tweets encoded using BERT, \texttt{bert-base-uncased}), in which users are surveyed for their age and gender \cite{cachola-etal-2018-expressively}. We bucket the age into three groups (0-25, 26-50 and above 50). Tweets in this dataset are annotated with their sentiment, ranging from one (very negative) to five (very positive). The dataset consists of more than 6,400 tweets written by more than 1,700 users. We removed users that no longer have public Twitter accounts and users with locations that do not exist based on a filter,\footnote{We used a list of cities, counties and states in the United States, taken from \url{https://tinyurl.com/4kmc6pyn}. All users were in the United States when the data was collected by the original curators.} resulting in a dataset with over 3,000 tweets, written by 817 unique users. As tweets are short by nature and their number is relatively small, the debiasing signal in this dataset, the amount of information it contains about the guarded attributes, might not be sufficient for the attribute removal. To amplify this signal, we concatenated each tweet in the dataset to at most ten other tweets from the same user.

We study the relationship between the main task of sentiment detection and the two protected attributes of age and gender. As a protected attribute $\myvec{z}$, we use the combination of both age and gender as a binary one-hot vector. This dataset presents a use-case for our algorithm of a composed protected attribute. Rather than using a classifier for predicting the sentiment, we use linear regression. Following \newcite{cachola-etal-2018-expressively}, we use Mean Absolute Error (MAE) to report the error of the sentiment predictions. Given that the sentiment is predicted as a continuous value, we cannot use the TPR gap as in previous sections. Rather, we use the following formula:

\begin{equation}
\mathrm{MAEGap} = \mathrm{std}(\mathrm{MAD}_{z=j} \mid j \in [m]), \label{eq:maegap}
\end{equation}

\noindent where $\mathrm{MAD}_{z=j} = \frac{1}{\ell} \sum_{i} | \eta_{ij} - \mu_j|$ where $i$ ranges over the set of size $\ell$ of examples with protected attribute value $j$, $\mu_j$ is the average of absolute $\rv{Y}$ prediction error for that set and $\eta_{ij}$ is the absolute difference between $\mu_j$ and the absolute error of example $i$.\footnote{The absolute error of prediction $a$ with true value $b$ is $|a-b|$.} The function $\mathrm{std}$ in this case indicates the standard deviation of the $m$ values of $\mathrm{MAD}_{z=j}$, $j \in [m]$.


\paragraph{Results}
Table~\ref{table:twitter} presents our results.
Overall, AMSAL reduces the gender and age gap in the predictions while not increasing by much MAE. In addition, we can see both AM-based methods outperform their $k$-means counterparts which increase unfairness (\textsc{Kmeans} + \textsc{INLP}) or significantly harm the downstream-task performance (\textsc{Kmeans} + \textsc{SAL}).
We also consider Figure~\ref{fig:twitter-age-gender}, which shows the quality of the assignments of the AM algorithm change as a function of the labeled data used. As expected, the more labeled data we have, the more accurate the assignments are, but the differences are not very large.

\begin{table}[]
\centering
\scalebox{0.7}{
\begin{tabular}{lHrHrHr}
\toprule
                    Model & mean squared error on Y & MAE & MSE age & Age (gap) & MSE gender & Gender (gap) \\
\toprule
                    \textsc{BertModel} &                     0.877 &                      0.745 &                0.558 &                0.031 &                   0.119 &                   0.011 \\
                  \, + \textsc{AM + INLP} &          \da{0.057} 0.821 &           \da{0.027} 0.717 &     \da{0.072} 0.485 &     \da{0.000} 0.031 &        \da{0.013} 0.106 &        \da{0.008} 0.003 \\
  \, + \textsc{Kmeans} + \textsc{INLP} &          \da{0.097} 0.780 &           \da{0.052} 0.693 &     \da{0.171} 0.386 &     \da{0.001} 0.030 &        \ua{0.067} 0.186 &        \ua{0.010} 0.021 \\
  \, + \textsc{Oracle}\textsc{INLP} &          \da{0.046} 0.831 &           \da{0.022} 0.723 &     \da{0.030} 0.527 &     \da{0.008} 0.022 &        \ua{0.093} 0.212 &        \ua{0.005} 0.017 \\
 \, + \textsc{Partial}\textsc{INLP} &          \da{0.048} 0.830 &           \da{0.025} 0.719 &     \da{0.130} 0.428 &     \ua{0.007} 0.038 &        \da{0.026} 0.093 &        \da{0.000} 0.011 \\
                   \, + \textsc{AMSAL} &          \ua{0.024} 0.901 &           \ua{0.009} 0.754 &     \da{0.020} 0.538 &     \da{0.005} 0.026 &        \da{0.043} 0.076 &        \da{0.009} 0.002 \\
   \, + \textsc{Kmeans} + \textsc{SAL} &          \ua{0.093} 0.970 &           \ua{0.039} 0.783 &     \ua{0.053} 0.610 &     \da{0.001} 0.030 &        \ua{0.003} 0.122 &        \da{0.007} 0.004 \\
   \, + \textsc{Oracle}\textsc{SAL} &          \ua{0.028} 0.905 &           \ua{0.012} 0.757 &     \ua{0.013} 0.570 &     \da{0.002} 0.029 &        \da{0.082} 0.037 &        \da{0.009} 0.003 \\
  \, + \textsc{Partial}\textsc{SAL} &          \ua{0.064} 0.941 &           \ua{0.025} 0.769 &     \ua{0.024} 0.582 &     \da{0.001} 0.030 &        \da{0.104} 0.015 &        \da{0.005} 0.006 \\

\bottomrule

\end{tabular}
}
\caption{MAE and debiasing gap values on the Twitter dataset, when using BERT to encode the tweets. For age and gender, we give the MAE gap as in Eq.~\refeq{eq:maegap}.\label{table:twitter}}
\end{table}

\begin{figure}[t]
    \centering
    \includegraphics[width=2.7in]{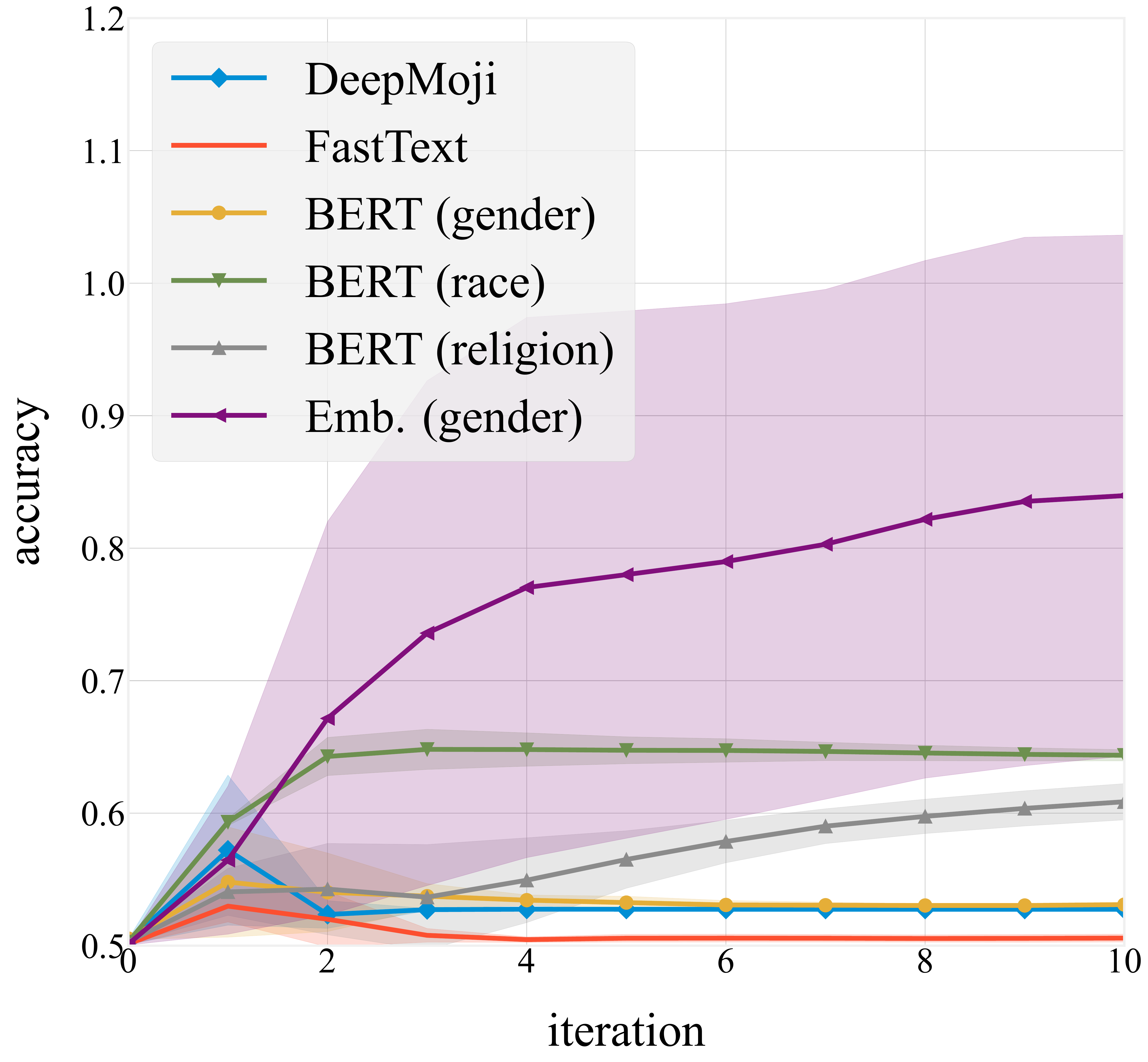}
    \caption{Accuracy of the AM steps (in identifying the correct assignment of inputs to guarded information) as a function of the iteration number. Shaded gray gives upper and lower bound on the standard deviation over five runs with different seeds for the initial $\pi$. FastText refers to the BiasBios dataset,  the BERT models are for the CrowS-Pairs dataset and Emb. refers to the word embeddings dataset from \S\ref{section:experiments-we}.}
    \label{fig:acc-stab}
\end{figure}

\begin{figure}[t]
    \centering
    \includegraphics[width=2.7in]{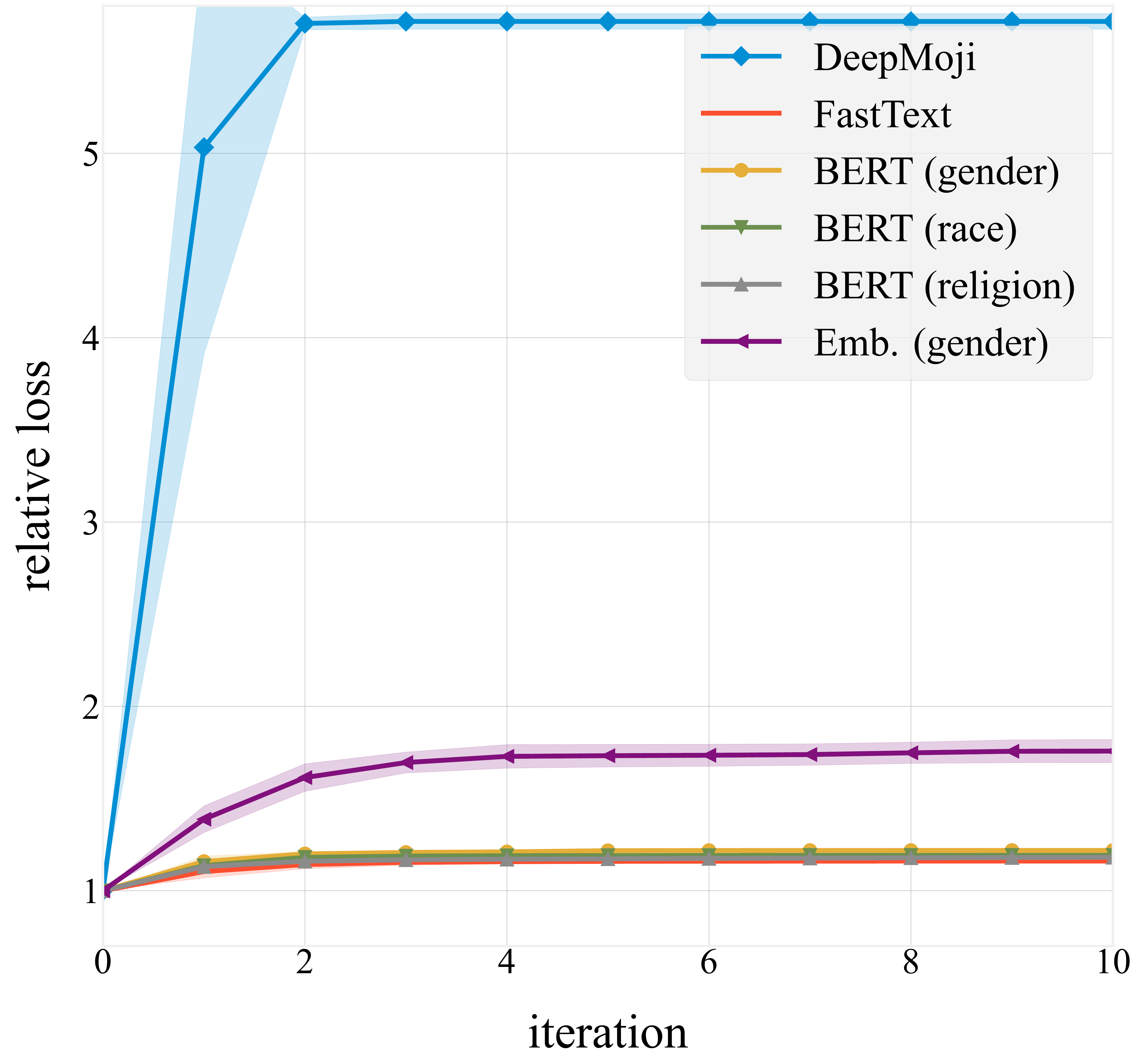}
    \caption{Ratio of the objective value in iteration $t$ and iteration $0$ of the ILP for the AM steps as a function of the iteration number $t$. Shaded gray gives upper and lower bound on the standard deviation over five runs with different seeds for the initial $\pi$. See legend explanation in Table~\ref{fig:acc-stab}.}
    \label{fig:svd}
\end{figure}

\begin{table*}[]

\begin{tabular}{lll}

\begin{minipage}{0.45\textwidth}
\scalebox{0.95}{

\begin{tabular}{lrr}
\toprule
                    Model & Task Acc. & TPR-GAP  \\
\midrule
                    \textsc{deepmoji} &             0.77 &                 0.14  \\
                 $\,$ + \textsc{AM + INLP} &  \uag{0.00} 0.77 &       \ua{0.00} 0.14  \\
 $\,$ + \textsc{Kmeans} + \textsc{INLP} &  \uag{0.00} 0.77 &       \ua{0.00} 0.14  \\
 $\,$ + \textsc{OracleINLP} &  \dab{0.02} 0.74 &       \da{0.04} 0.10  \\
$\,$ + \textsc{PartialINLP} &  \dab{0.01} 0.75 &       \da{0.06} 0.08  \\
                  $\,$ + \textsc{AMSAL} &  \dab{0.24} 0.52 &       \ua{0.03} 0.17  \\
  $\,$ + \textsc{Kmeans} + \textsc{SAL} &  \dab{0.23} 0.54 &       \ua{0.12} 0.26  \\
  $\,$ + \textsc{OracleSAL} &  \dab{0.00} 0.76 &       \da{0.03} 0.11  \\
 $\,$ + \textsc{PartialSAL} &  \dab{0.19} 0.57 &       \ua{0.15} 0.29  \\

\bottomrule
\end{tabular}
}
\end{minipage}

 & &

\begin{minipage}{0.5\textwidth}

\centering
\scalebox{0.95}{
\begin{tabular}{lrr}
\toprule
                    Model &   $F_1$ (macro) & TPR-GAP  \\
\midrule
                    \textsc{deepmoji} &  0.66 &               0.06                    \\
                 $\,$ + \textsc{AM + INLP} &   \dab{0.0002} 0.66 &       \da{0.00} 0.06  \\
 $\,$ + \textsc{Kmeans} + \textsc{INLP} &     \dab{0.1} 0.56 &       \da{0.02} 0.04  \\
 $\,$ + \textsc{OracleINLP}  &   \dab{0.19} 0.46 &       \da{0.06} 0.00 \\
$\,$ + \textsc{PartialINLP}   &   \dab{0.14} 0.52 &       \da{0.03} 0.03 \\
                  $\,$ + \textsc{AMSAL}   &   \dab{0.16} 0.49 &       \da{0.02} 0.04\\
  $\,$ + \textsc{Kmeans} + \textsc{SAL}   &   \dab{0.17} 0.48 &       \da{0.04} 0.02 \\
  $\,$ + \textsc{OracleSAL}  &       \dab{0.01} 0.65  & \ua{0.03} 0.09  \\
 $\,$ + \textsc{PartialSAL}  &        \dab{0.11} 0.54 & \ua{0.00} 0.06  \\
\bottomrule
\end{tabular}
}

\end{minipage}

\end{tabular}

\caption{The performance of removing race information from the DeepMoji dataset is shown for two cases: with balanced ratios of race and sentiment (left) and with ratios of 0.8 for sentiment and 0.5 for race (right). In both cases, the total size of the dataset used is 30,000 examples. To evaluate the performance of the unbalanced sentiment dataset, we use the $F_1$ macro measure, because in an unbalanced dataset such as this one, a simple classifier that always returns one label will achieve an accuracy of 80\%. Such a classifier would have a $F_1$ macro score of $0.44\dot{4}$. \label{table:sentiment}}

\end{table*}

\subsection{An Example of Our Method Limitations}
\label{section:limitations}

We now present the main limitation in our approach and setting. This limitation arises when the random variables $\rv{Y}$ and $\rv{Z}$ are not easily distinguishable through information about $\rv{X}$.

We experiment with a binary sentiment analysis ($\myvec{y}$) task, predicted on users' tweets ($\myvec{x}$), aiming to remove information regarding the authors' ethnic affiliations. To do so, we use a dataset collected by \newcite{blodgett2016demographic}, which examined the differences between African-American English (AAE) speakers and Standard American English (SAE) speakers. As information about one's ethnicity is hard to obtain, the user's geolocation information was used to create a distantly supervised mapping between authors and their ethnic affiliations. We follow previous work \cite{shao2022gold, ravfogel2020null} and use the DeepMoji encoder \cite{felbo2017using} to obtain representations for the tweets. The train and test sets are balanced regarding sentiment and authors' ethnicity. 
We use 20\% of the examples for the \textsc{Partial} setting. Table~\ref{table:sentiment} gives the results for this dataset. We observe that the removal with the assignment ($k$-means, AM or \textsc{Partial}) significantly harms the performance on the main task and reduces it to a random guess.

This presents a limitation of our algorithm. A priori, there is no distinction between $\rv{Y}$ and $\rv{Z}$, as our method is unsupervised. In addition, the positive labels of $\rv{Y}$ and $\rv{Z}$ have the same prior probability. Indeed, when we check the assignment accuracy in the sentiment dataset, we observe that the $k$-means, AM and \textsc{Partial} AM assignment accuracy for identifying $\rv{Z}$ are between 0.55 and 0.59. If we check the assignment against $\rv{Y}$, we get an accuracy between 0.74 and 0.76. This means that all assignment algorithms actually identify $\rv{Y}$ rather than $\rv{Z}$ (both $\rv{Y}$ and $\rv{Z}$ are binary variables in this case). The conclusion from this is that our algorithm works best when sufficient information on $\rv{Z}$ is presented such that it can provide a basis for aligning samples of $\rv{Z}$ with samples of $\rv{X}$. Suppose such information is unavailable or unidentifiable with information regarding $\rv{Y}$. In that case, we may simply identify the natural clustering of $\rv{X}$ according to their main task classes, leading to low main-task performance.

In Table~\ref{table:sentiment}, we observe that this behavior is significantly mitigated when the priors over the sentiment and the race are different (0.8 for sentiment and 0.5 for race). In that case, the AM algorithm is able to distinguish between the race-protected attribute ($\myvec{z}$) and the sentiment class ($\myvec{y}$) quite consistently with INLP and SAL, and the gap is reduced.

We also observe that INLP changed neither the accuracy nor the TPR-GAP for the balanced scenario (Table~\ref{table:sentiment}) when using a $k$-means assignment or an AM assignment. Upon inspection, we found out that INLP returns an identity projection in these cases, unable to amplify the relatively weak signal in the assignment to change the representations.

\subsection{Stability Analysis of the Alignment}
\label{section:stability}

In Figure~\ref{fig:acc-stab}, we plot the accuracy of the alignment algorithm (knowing the true value of the guarded attribute per input) throughout the execution of the AM steps for the first ten iterations. The shaded area indicates one standard deviation. We observe that the first few iterations are the ones in which the accuracy improves the most. For most of the datasets, the accuracy does not decrease between iterations, though in the case of DeepMoji we do observe a ``bump.'' This is indeed why the \textsc{Partial} setting of our algorithm, where a small amount of guarded information is available to determine at which iteration to stop the AM algorithm, is important.  In the word embeddings case, the variance is larger because, in certain executions, the algorithm converged quickly, while in others, it took more iterations to converge to high accuracy.

Figure~\ref{fig:svd} plots the relative change of the objective value of the ILP from \S\ref{section:ilp} against iteration number. The relative change is defined as the ratio between the objective value before the algorithm begins and the same value at a given iteration. We see that there is a relative stability of the algorithm and that the AM steps converge quite quickly. We also observe the DeepMoji dataset has a large increase in the objective value in the first iteration (around $\times 5$ compared to the value the algorithm starts with), after which it remains stable.



\section{Related Work}



There has been an increasing amount of work about detecting and erasing undesired or protected information from neural representations, with standard software packages for this process having been developed \cite{han2022fairlib}. For example, in their seminal work, \newcite{bolukbasi2016man} showed that word embeddings exhibit gender stereotypes. To mitigate this issue, they projected the word embeddings to a neutral space with respect to a ``he-she'' direction. Influenced by this work, \newcite{zhao-etal-2018-learning} proposed a customized training scheme to reduce the gender bias in word embeddings. \newcite{gonen2019lipstick} examined the effectiveness of the methods mentioned above and concluded they remove bias in a shallow way. For example, they demonstrated that classifiers can accurately predict the gender associated with a  word when fed with the embeddings of both debiasing methods.

Another related strand of work uses adversarial learning \cite{ganin2016domain}, where an additional objective function is added for balancing undesired-information removal and the main task \cite{edwards2015censoring,li2018towards,coavoux2018privacy, wang2021dynamically}.  \newcite{elazar2018adversarial} have also demonstrated that an ad-hoc classifier can easily recover the removed information from adversarially trained representations.
Since then, methods for information erasure such as INLP and its generalization \cite{ravfogel2020null,ravfogel2022linear}, SAL \cite{shao2022gold} and methods based on similarity measures between neural representations \cite{colombo2022learning} have been developed. With a similar motivation to ours, \newcite{han2021decoupling} aimed to ease the burden of obtaining guarded attributes at a large scale by decoupling the adversarial information removal process from the main task training. They, however, did not experiment with debiasing representations where no guarded attribute alignments are available. \newcite{shao2022gold} experimented with the removal of features in a scenario in which a low number of protected attributes is available.

Additional previous work showed that methods based on causal inference \cite{feder2021causalm}, train-set balancing \cite{han2021balancing}, and contrastive learning \cite{shen2021contrastive,chi2022conditional} effectively reduce bias and increase fairness.  In addition, there is a large body of work for detecting bias, its evaluation \citep{dev-etal-2021-oscar}
and its implications in specific NLP applications. \newcite{savoldi-etal-2022-morphosyntactic} detected a gender bias in speech translation systems for gendered languages. Gender bias is also discussed in the context of knowledge base embeddings by \newcite{fisher2019measuring, du2022understanding}, and multilingual text classification \cite{huang2022easy}. 






\section{Conclusions and Future Work}
We presented a new and challenging setup for removing information, with minimal or no available sensitive information alignment. This setup is crucial for the wide applicability of debiasing methods, as for most applications, obtaining such sensitive labels on a large scale is challenging. To ease this problem, we present a method to erase information from neural representations, where the guarded attribute information does not accompany each input instance. Our main algorithm, AMSAL, alternates between two steps (Assignment and Maximization) to identify an assignment between the input instances and the guarded information records. It then completes its execution by removing the information by minimizing covariance between the input instances and the aligned guarded attributes. Our approach is modular, and other erasure algorithms, such as INLP, can be used with it. Experiments show that we can reduce the unwanted bias in many cases while keeping the representations highly useful. Future work might include extending our technique to the kernelized case, analogously to the method of \newcite{shao2022gold}.


\ignore{
\section{Limitations}
\label{section:limits}
Our method works under the following assumptions:

\begin{itemize}
    
    \item The distribution over Z is known. 
    
\end{itemize}

don't recommend to use it to align overall, but rather it works when aggregated removal happens, not through $\pi$ directly.
}

\section*{Ethical Considerations}

The AM algorithm could potentially be misused by rather than using the AM steps to erase information, using them to link records of two different types, undermining the privacy of the record holders. Such a situation may merit additional concern because the links returned between the guarded attributes and the input instances will likely contain mistakes. The links are unreliable for decision-making at the \emph{individual level}. Instead, they should be used on an aggregate as a statistical construct to erase information from the input representations.
Finally,\footnote{We thank the anonymous reviewer for raising this issue.} we note that the automation of the debiasing process, without properly statistically confirming its accuracy using a correct sample may promote a false sense of security that a given system is making fair decisions. We do not recommend using our method for debiasing without proper statistical control and empirical verification of correctness.



\section*{Acknowledgments}

We thank the reviewers, the action editors and Marcio Fonseca for their thorough feedback. We also thank Daniel Preoțiuc-Pietro for his help with the Twitter data. We thank Kousha Etessami for being a sounding board for certain parts of the paper.
The experiments in this paper were supported by compute grants from the Edinburgh Parallel Computing Center and from the Baskerville Tier 2 HPC service (University of Birmingham).

\ignore{
\section{Shun's Observations}
\begin{itemize}
    \item We are more stable compared to the clusters methods when it comes to different sub sets
    \item Shun's first observations 
    \item Shun's second observations 
\end{itemize}
}

\bibliography{anthology,custom}

\begin{thebibliography}{39}
\expandafter\ifx\csname natexlab\endcsname\relax\def\natexlab#1{#1}\fi

\bibitem[{Blodgett et~al.(2016)Blodgett, Green, and
  O{'}Connor}]{blodgett2016demographic}
Su~Lin Blodgett, Lisa Green, and Brendan O{'}Connor. 2016.
\newblock \href {https://doi.org/10.18653/v1/D16-1120} {Demographic dialectal
  variation in social media: A case study of {A}frican-{A}merican {E}nglish}.
\newblock In \emph{Proceedings of the 2016 Conference on Empirical Methods in
  Natural Language Processing}, pages 1119--1130, Austin, Texas. Association
  for Computational Linguistics.

\bibitem[{Bolukbasi et~al.(2016)Bolukbasi, Chang, Zou, Saligrama, and
  Kalai}]{bolukbasi2016man}
Tolga Bolukbasi, Kai{-}Wei Chang, James~Y. Zou, Venkatesh Saligrama, and
  Adam~Tauman Kalai. 2016.
\newblock \href
  {https://proceedings.neurips.cc/paper/2016/hash/a486cd07e4ac3d270571622f4f316ec5-Abstract.html}
  {Man is to computer programmer as woman is to homemaker? debiasing word
  embeddings}.
\newblock In \emph{Advances in Neural Information Processing Systems 29: Annual
  Conference on Neural Information Processing Systems 2016, December 5-10,
  2016, Barcelona, Spain}, pages 4349--4357.

\bibitem[{Cachola et~al.(2018)Cachola, Holgate, Preo{\c{t}}iuc-Pietro, and
  Li}]{cachola-etal-2018-expressively}
Isabel Cachola, Eric Holgate, Daniel Preo{\c{t}}iuc-Pietro, and Junyi~Jessy Li.
  2018.
\newblock \href {https://aclanthology.org/C18-1248} {Expressively vulgar: The
  socio-dynamics of vulgarity and its effects on sentiment analysis in social
  media}.
\newblock In \emph{Proceedings of the 27th International Conference on
  Computational Linguistics}, pages 2927--2938, Santa Fe, New Mexico, USA.
  Association for Computational Linguistics.

\bibitem[{Caliskan et~al.(2017)Caliskan, Bryson, and
  Narayanan}]{caliskan2017semantics}
Aylin Caliskan, Joanna~J Bryson, and Arvind Narayanan. 2017.
\newblock Semantics derived automatically from language corpora contain
  human-like biases.
\newblock \emph{Science}, 356(6334):183--186.

\bibitem[{Chi et~al.(2022)Chi, Shand, Yu, Chang, Zhao, and
  Tian}]{chi2022conditional}
Jianfeng Chi, William Shand, Yaodong Yu, Kai-Wei Chang, Han Zhao, and Yuan
  Tian. 2022.
\newblock \href {https://arxiv.org/abs/2205.11485} {Conditional supervised
  contrastive learning for fair text classification}.
\newblock \emph{ArXiv preprint}, abs/2205.11485.

\bibitem[{Coavoux et~al.(2018)Coavoux, Narayan, and Cohen}]{coavoux2018privacy}
Maximin Coavoux, Shashi Narayan, and Shay~B. Cohen. 2018.
\newblock \href {https://doi.org/10.18653/v1/D18-1001} {Privacy-preserving
  neural representations of text}.
\newblock In \emph{Proceedings of the 2018 Conference on Empirical Methods in
  Natural Language Processing}, pages 1--10, Brussels, Belgium. Association for
  Computational Linguistics.

\bibitem[{Colombo et~al.(2022)Colombo, Staerman, Noiry, and
  Piantanida}]{colombo2022learning}
Pierre Colombo, Guillaume Staerman, Nathan Noiry, and Pablo Piantanida. 2022.
\newblock \href {https://doi.org/10.18653/v1/2022.acl-long.187} {Learning
  disentangled textual representations via statistical measures of similarity}.
\newblock In \emph{Proceedings of the 60th Annual Meeting of the Association
  for Computational Linguistics (Volume 1: Long Papers)}, pages 2614--2630,
  Dublin, Ireland. Association for Computational Linguistics.

\bibitem[{De-Arteaga et~al.(2019)De-Arteaga, Romanov, Wallach, Chayes, Borgs,
  Chouldechova, Geyik, Kenthapadi, and Kalai}]{de2019bias}
Maria De-Arteaga, Alexey Romanov, Hanna Wallach, Jennifer Chayes, Christian
  Borgs, Alexandra Chouldechova, Sahin Geyik, Krishnaram Kenthapadi, and
  Adam~Tauman Kalai. 2019.
\newblock Bias in bios: A case study of semantic representation bias in a
  high-stakes setting.
\newblock In \emph{proceedings of the Conference on Fairness, Accountability,
  and Transparency}, pages 120--128.

\bibitem[{Devlin et~al.(2019)Devlin, Chang, Lee, and
  Toutanova}]{devlin2018bert}
Jacob Devlin, Ming-Wei Chang, Kenton Lee, and Kristina Toutanova. 2019.
\newblock \href {https://doi.org/10.18653/v1/N19-1423} {{BERT}: Pre-training of
  deep bidirectional transformers for language understanding}.
\newblock In \emph{Proceedings of the 2019 Conference of the North {A}merican
  Chapter of the Association for Computational Linguistics: Human Language
  Technologies, Volume 1 (Long and Short Papers)}, pages 4171--4186,
  Minneapolis, Minnesota. Association for Computational Linguistics.

\bibitem[{Du et~al.(2022)Du, Zheng, Wu, Lan, Yang, and
  Ma}]{du2022understanding}
Yupei Du, Qi~Zheng, Yuanbin Wu, Man Lan, Yan Yang, and Meirong Ma. 2022.
\newblock \href {https://doi.org/10.18653/v1/2022.acl-long.98} {Understanding
  gender bias in knowledge base embeddings}.
\newblock In \emph{Proceedings of the 60th Annual Meeting of the Association
  for Computational Linguistics (Volume 1: Long Papers)}, pages 1381--1395,
  Dublin, Ireland. Association for Computational Linguistics.

\bibitem[{Edwards and Storkey(2016)}]{edwards2015censoring}
Harrison Edwards and Amos~J. Storkey. 2016.
\newblock \href {http://arxiv.org/abs/1511.05897} {Censoring representations
  with an adversary}.
\newblock In \emph{4th International Conference on Learning Representations,
  {ICLR} 2016, San Juan, Puerto Rico, May 2-4, 2016, Conference Track
  Proceedings}.

\bibitem[{Elazar and Goldberg(2018)}]{elazar2018adversarial}
Yanai Elazar and Yoav Goldberg. 2018.
\newblock \href {https://doi.org/10.18653/v1/D18-1002} {Adversarial removal of
  demographic attributes from text data}.
\newblock In \emph{Proceedings of the 2018 Conference on Empirical Methods in
  Natural Language Processing}, pages 11--21, Brussels, Belgium. Association
  for Computational Linguistics.

\bibitem[{Feder et~al.(2021)Feder, Oved, Shalit, and
  Reichart}]{feder2021causalm}
Amir Feder, Nadav Oved, Uri Shalit, and Roi Reichart. 2021.
\newblock \href {https://doi.org/10.1162/coli_a_00404} {{C}ausa{LM}: Causal
  model explanation through counterfactual language models}.
\newblock \emph{Computational Linguistics}, 47(2):333--386.

\bibitem[{Felbo et~al.(2017)Felbo, Mislove, S{\o}gaard, Rahwan, and
  Lehmann}]{felbo2017using}
Bjarke Felbo, Alan Mislove, Anders S{\o}gaard, Iyad Rahwan, and Sune Lehmann.
  2017.
\newblock \href {https://doi.org/10.18653/v1/D17-1169} {Using millions of emoji
  occurrences to learn any-domain representations for detecting sentiment,
  emotion and sarcasm}.
\newblock In \emph{Proceedings of the 2017 Conference on Empirical Methods in
  Natural Language Processing}, pages 1615--1625, Copenhagen, Denmark.
  Association for Computational Linguistics.

\bibitem[{Fisher et~al.(2019)Fisher, Palfrey, Christodoulopoulos, and
  Mittal}]{fisher2019measuring}
Joseph Fisher, Dave Palfrey, Christos Christodoulopoulos, and Arpit Mittal.
  2019.
\newblock \href {https://arxiv.org/abs/1912.02761} {Measuring social bias in
  knowledge graph embeddings}.
\newblock \emph{ArXiv preprint}, abs/1912.02761.

\bibitem[{Ganin et~al.(2016)Ganin, Ustinova, Ajakan, Germain, Larochelle,
  Laviolette, Marchand, and Lempitsky}]{ganin2016domain}
Yaroslav Ganin, Evgeniya Ustinova, Hana Ajakan, Pascal Germain, Hugo
  Larochelle, Fran{\c{c}}ois Laviolette, Mario Marchand, and Victor Lempitsky.
  2016.
\newblock Domain-adversarial training of neural networks.
\newblock \emph{The journal of machine learning research}, 17(1):2096--2030.

\bibitem[{Gonen and Goldberg(2019)}]{gonen2019lipstick}
Hila Gonen and Yoav Goldberg. 2019.
\newblock \href {https://doi.org/10.18653/v1/N19-1061} {Lipstick on a pig:
  {D}ebiasing methods cover up systematic gender biases in word embeddings but
  do not remove them}.
\newblock In \emph{Proceedings of the 2019 Conference of the North {A}merican
  Chapter of the Association for Computational Linguistics: Human Language
  Technologies, Volume 1 (Long and Short Papers)}, pages 609--614, Minneapolis,
  Minnesota. Association for Computational Linguistics.

\bibitem[{Han et~al.(2021{\natexlab{a}})Han, Baldwin, and
  Cohn}]{han2021balancing}
Xudong Han, Timothy Baldwin, and Trevor Cohn. 2021{\natexlab{a}}.
\newblock \href {https://arxiv.org/abs/2109.08253} {Balancing out bias:
  Achieving fairness through training reweighting}.
\newblock \emph{ArXiv preprint}, abs/2109.08253.

\bibitem[{Han et~al.(2021{\natexlab{b}})Han, Baldwin, and
  Cohn}]{han2021decoupling}
Xudong Han, Timothy Baldwin, and Trevor Cohn. 2021{\natexlab{b}}.
\newblock \href {https://doi.org/10.18653/v1/2021.findings-acl.41} {Decoupling
  adversarial training for fair {NLP}}.
\newblock In \emph{Findings of the Association for Computational Linguistics:
  ACL-IJCNLP 2021}, pages 471--477, Online. Association for Computational
  Linguistics.

\bibitem[{Han et~al.(2022)Han, Shen, Li, Frermann, Baldwin, and
  Cohn}]{han2022fairlib}
Xudong Han, Aili Shen, Yitong Li, Lea Frermann, Timothy Baldwin, and Trevor
  Cohn. 2022.
\newblock \href {https://arxiv.org/abs/2205.01876} {fairlib: A unified
  framework for assessing and improving classification fairness}.
\newblock \emph{ArXiv preprint}, abs/2205.01876.

\bibitem[{Huang(2022)}]{huang2022easy}
Xiaolei Huang. 2022.
\newblock \href {https://doi.org/10.18653/v1/2022.naacl-main.52} {Easy
  adaptation to mitigate gender bias in multilingual text classification}.
\newblock In \emph{Proceedings of the 2022 Conference of the North American
  Chapter of the Association for Computational Linguistics: Human Language
  Technologies}, pages 717--723, Seattle, United States. Association for
  Computational Linguistics.

\bibitem[{Joulin et~al.(2016)Joulin, Grave, Bojanowski, Douze, J{\'e}gou, and
  Mikolov}]{joulin2016fasttext}
Armand Joulin, Edouard Grave, Piotr Bojanowski, Matthijs Douze, H{\'e}rve
  J{\'e}gou, and Tomas Mikolov. 2016.
\newblock \href {https://arxiv.org/abs/1612.03651} {Fasttext. zip: Compressing
  text classification models}.
\newblock \emph{ArXiv preprint}, abs/1612.03651.

\bibitem[{Kuhn(1955)}]{Kuhn1955}
Harold~W. Kuhn. 1955.
\newblock The hungarian method for the assignment problem.
\newblock \emph{Naval Research Logistics Quarterly}, 2:83--97.

\bibitem[{Li et~al.(2018)Li, Baldwin, and Cohn}]{li2018towards}
Yitong Li, Timothy Baldwin, and Trevor Cohn. 2018.
\newblock \href {https://doi.org/10.18653/v1/P18-2005} {Towards robust and
  privacy-preserving text representations}.
\newblock In \emph{Proceedings of the 56th Annual Meeting of the Association
  for Computational Linguistics (Volume 2: Short Papers)}, pages 25--30,
  Melbourne, Australia. Association for Computational Linguistics.

\bibitem[{MacKay(2003)}]{mackay2003information}
David J~C MacKay. 2003.
\newblock \emph{Information theory, inference and learning algorithms}.
\newblock Cambridge university press.

\bibitem[{May et~al.(2019)May, Wang, Bordia, Bowman, and
  Rudinger}]{may-etal-2019-measuring}
Chandler May, Alex Wang, Shikha Bordia, Samuel~R. Bowman, and Rachel Rudinger.
  2019.
\newblock \href {https://doi.org/10.18653/v1/N19-1063} {On measuring social
  biases in sentence encoders}.
\newblock In \emph{Proceedings of the 2019 Conference of the North {A}merican
  Chapter of the Association for Computational Linguistics: Human Language
  Technologies, Volume 1 (Long and Short Papers)}, pages 622--628, Minneapolis,
  Minnesota. Association for Computational Linguistics.

\bibitem[{Meade et~al.(2022)Meade, Poole-Dayan, and Reddy}]{meade2022empirical}
Nicholas Meade, Elinor Poole-Dayan, and Siva Reddy. 2022.
\newblock \href {https://doi.org/10.18653/v1/2022.acl-long.132} {An empirical
  survey of the effectiveness of debiasing techniques for pre-trained language
  models}.
\newblock In \emph{Proceedings of the 60th Annual Meeting of the Association
  for Computational Linguistics (Volume 1: Long Papers)}, pages 1878--1898,
  Dublin, Ireland. Association for Computational Linguistics.

\bibitem[{Nadeem et~al.(2021)Nadeem, Bethke, and
  Reddy}]{nadeem-etal-2021-stereoset}
Moin Nadeem, Anna Bethke, and Siva Reddy. 2021.
\newblock \href {https://doi.org/10.18653/v1/2021.acl-long.416} {{S}tereo{S}et:
  Measuring stereotypical bias in pretrained language models}.
\newblock In \emph{Proceedings of the 59th Annual Meeting of the Association
  for Computational Linguistics and the 11th International Joint Conference on
  Natural Language Processing (Volume 1: Long Papers)}, pages 5356--5371,
  Online. Association for Computational Linguistics.

\bibitem[{Nangia et~al.(2020)Nangia, Vania, Bhalerao, and
  Bowman}]{nangia2020crows}
Nikita Nangia, Clara Vania, Rasika Bhalerao, and Samuel~R. Bowman. 2020.
\newblock \href {https://doi.org/10.18653/v1/2020.emnlp-main.154}
  {{C}row{S}-pairs: A challenge dataset for measuring social biases in masked
  language models}.
\newblock In \emph{Proceedings of the 2020 Conference on Empirical Methods in
  Natural Language Processing (EMNLP)}, pages 1953--1967, Online. Association
  for Computational Linguistics.

\bibitem[{Ramshaw and Tarjan(2012)}]{ramshaw2012minimum}
Lyle Ramshaw and Robert~E Tarjan. 2012.
\newblock On minimum-cost assignments in unbalanced bipartite graphs.
\newblock \emph{HP Labs, Palo Alto, CA, USA, Tech. Rep. HPL-2012-40R1}.

\bibitem[{Ravfogel et~al.(2020)Ravfogel, Elazar, Gonen, Twiton, and
  Goldberg}]{ravfogel2020null}
Shauli Ravfogel, Yanai Elazar, Hila Gonen, Michael Twiton, and Yoav Goldberg.
  2020.
\newblock \href {https://doi.org/10.18653/v1/2020.acl-main.647} {Null it out:
  Guarding protected attributes by iterative nullspace projection}.
\newblock In \emph{Proceedings of the 58th Annual Meeting of the Association
  for Computational Linguistics}, pages 7237--7256, Online. Association for
  Computational Linguistics.

\bibitem[{Ravfogel et~al.(2022)Ravfogel, Twiton, Goldberg, and
  Cotterell}]{ravfogel2022linear}
Shauli Ravfogel, Michael Twiton, Yoav Goldberg, and Ryan~D Cotterell. 2022.
\newblock Linear adversarial concept erasure.
\newblock In \emph{International Conference on Machine Learning}, pages
  18400--18421. PMLR.

\bibitem[{Savoldi et~al.(2022)Savoldi, Gaido, Bentivogli, Negri, and
  Turchi}]{savoldi-etal-2022-morphosyntactic}
Beatrice Savoldi, Marco Gaido, Luisa Bentivogli, Matteo Negri, and Marco
  Turchi. 2022.
\newblock \href {https://doi.org/10.18653/v1/2022.acl-long.127} {Under the
  morphosyntactic lens: A multifaceted evaluation of gender bias in speech
  translation}.
\newblock In \emph{Proceedings of the 60th Annual Meeting of the Association
  for Computational Linguistics (Volume 1: Long Papers)}, pages 1807--1824,
  Dublin, Ireland. Association for Computational Linguistics.

\bibitem[{Shao et~al.(2023)Shao, Ziser, and Cohen}]{shao2022gold}
Shun Shao, Yftah Ziser, and Shay~B Cohen. 2023.
\newblock \href {https://arxiv.org/abs/2203.07893} {Gold doesn't always
  glitter: Spectral removal of linear and nonlinear guarded attribute
  information}.
\newblock In \emph{Proceedings of the 17th Annual Meeting of the European
  chapter of the Association for Computational Linguistics (EACL)}, volume
  abs/2203.07893.

\bibitem[{Shen et~al.(2021)Shen, Han, Cohn, Baldwin, and
  Frermann}]{shen2021contrastive}
Aili Shen, Xudong Han, Trevor Cohn, Timothy Baldwin, and Lea Frermann. 2021.
\newblock \href {https://arxiv.org/abs/2109.10645} {Contrastive learning for
  fair representations}.
\newblock \emph{ArXiv preprint}, abs/2109.10645.

\bibitem[{Stewart(1990)}]{stewart1998perturbation}
Gilbert~W Stewart. 1990.
\newblock Perturbation theory for the singular value decomposition.
\newblock Technical Report UMIACS-90-120 / CS-TR 2539, University of Maryland,
  College Park.

\bibitem[{Wang et~al.(2019)Wang, Singh, Michael, Hill, Levy, and
  Bowman}]{wang2018glue}
Alex Wang, Amanpreet Singh, Julian Michael, Felix Hill, Omer Levy, and
  Samuel~R. Bowman. 2019.
\newblock \href {https://openreview.net/forum?id=rJ4km2R5t7} {{GLUE:} {A}
  multi-task benchmark and analysis platform for natural language
  understanding}.
\newblock In \emph{7th International Conference on Learning Representations,
  {ICLR} 2019, New Orleans, LA, USA, May 6-9, 2019}. OpenReview.net.

\bibitem[{Wang et~al.(2021)Wang, Yan, He, Wu, and Xu}]{wang2021dynamically}
Liwen Wang, Yuanmeng Yan, Keqing He, Yanan Wu, and Weiran Xu. 2021.
\newblock \href {https://doi.org/10.18653/v1/2021.naacl-main.293} {Dynamically
  disentangling social bias from task-oriented representations with adversarial
  attack}.
\newblock In \emph{Proceedings of the 2021 Conference of the North American
  Chapter of the Association for Computational Linguistics: Human Language
  Technologies}, pages 3740--3750, Online. Association for Computational
  Linguistics.

\bibitem[{Zhao et~al.(2018)Zhao, Zhou, Li, Wang, and
  Chang}]{zhao-etal-2018-learning}
Jieyu Zhao, Yichao Zhou, Zeyu Li, Wei Wang, and Kai-Wei Chang. 2018.
\newblock \href {https://doi.org/10.18653/v1/D18-1521} {Learning gender-neutral
  word embeddings}.
\newblock In \emph{Proceedings of the 2018 Conference on Empirical Methods in
  Natural Language Processing}, pages 4847--4853, Brussels, Belgium.
  Association for Computational Linguistics.

\end{thebibliography}
\bibliographystyle{acl_natbib}


\onlywithappendix{

\appendix

\section{Justification of the AM Algorithm: Further Details}
\label{appendix:just}

We provide here the full details for the claim in \S\ref{section:justification}.
Our first observation is that for a uniformly sampled permutation $\pi \colon [n] \rightarrow [n]$, the probability that it has exactly $k \le n$ elements such that $\pi(i) = i$ for all $i$ in this set of elements is bounded
from above by:\footnote{Choose $k$ elements that are fixed, and let the rest vary arbitrarily.}

\begin{equation}
    \displaystyle\frac{\displaystyle{\binom{n}{k}}(n-k)!}{n!} = \displaystyle\frac{1}{k!}.
\end{equation}


We also assume that $\mathbb{E}[\rv{X} \mid \rv{H}] = 0$ and $\mathbb{E}[\rv{Z} \mid \rv{H}] = 0$, and that the product of every pair of coordinates of $\rv{X}$ and $\rv{Z}$ is
bounded in absolute value by a constant $B > 0$.
Let $\{ (\myvec{x}^{(i)}, \myvec{z}^{(i)}, \myvec{h}^{(i)} ) \}$ be a random sample of size $n$ from the joint distribution $p(\rv{X}, \rv{Z}, \rv{H})$.
Given a permutation $\pi \colon [n] \rightarrow [n]$, define $I(\pi) = \{ i \mid \pi(i) = i \}$. For a given set $M \subseteq [n]$, define

\begin{equation}
\mymat{\Omega}_{\pi \mid M} = \sum_{i \in M} \myvec{x}^{(i)} (\myvec{z}^{(\pi(i))})^{\top}.
\end{equation}

For a matrix $\mymat{A} \in \mathbb{R}^{d \times d'}$, let $\sigma_j(\mymat{A})$ be its $j$th largest singular value, and
let $\sigma^+(\mymat{A}) = \sum_j \sigma_j(\mymat{A})$. Let $\sigma^+ = \sigma^+(\mathbb{E}[\mymat{\Omega}_{\iota}])$.

We first note that for any permutation $\pi$, it holds that $\mathbb{E}[\Omega_{\pi \mid K}] = 0$ where we define
$K = [n] \setminus I(\pi)$.

\begin{lemma}\label{lemma:hoeff}
For any $t > 0$, it holds that:

\begin{equation}
p(||\mymat{\Omega}_{\pi \mid I(\pi)} - \mathbb{E}[\mymat{\Omega}_{\pi \mid I(\pi)}]||_2 \ge dd't)
\label{eq:omega-pi}
\end{equation}

\noindent is smaller than $2dd' \exp \left(- \displaystyle\frac{t^2}{|I(\pi)|B^2} \right).$

\end{lemma}

\begin{proof}
By Hoeffding's inequality, for any $i \in [d]$, $j \in [d']$, it holds that the probability
that for $|I(\pi)|$ i.i.d. r.v. $\rv{X}^k$, $\rv{Z}^k$ the following is true:

\begin{align}
\left|\sum_{k \in I(\pi)} \elementrv{X}^k_i \elementrv{Z}^k_j - \sum_{k \in I(\pi)} \mathbb{E}[\elementrv{X}^k_i \elementrv{Z}^k_j]\right| \ge t
\end{align}

\noindent is smaller than $2 \exp \left(- \displaystyle\frac{t^2}{|I(\pi)|B^2} \right).$
Therefore, by a union bound on each element of the matrix $\Omega_{\pi}$, we get the upper bound on
Eq.~\refeq{eq:omega-pi}.

\end{proof}

\begin{lemma}\label{lem:simple}
For any $t > 0$, it holds that:

\begin{equation}
||\mymat{\Omega}_{\pi \mid K} - \mathbb{E}[\mymat{\Omega}_{\pi \mid K}]||_2
\label{eq:omega-pi2}
\end{equation}

\noindent is smaller than $2|K|dd'B$.

\end{lemma}

\begin{proof}
Since $\elementrv{X}_i$ and $\elementrv{Z}_j$ are bounded as a product in absolute value by $B$,
and the dimensions of $\mymat{\Omega}_{\pi \mid K}$ is $d \times d'$, each cell being a sum of $|K|$
values, the bound naturally follows.
\end{proof}

Let $n$ such that $n \sigma^+ > 2kdd'B$ where $k = |K|$.
Then from Lemma~\ref{lem:simple}, $||\mymat{\Omega}_{\pi \mid K} - \mathbb{E}[\mymat{\Omega}_{\pi \mid K}]||_2 < n\sigma^+$.
Consider the event $\sigma^+(\mymat{\Omega}_{\iota}) < \sigma^+(\mymat{\Omega}_{\pi})$.
Its probability is bounded from above by the probability of the event $\sigma^+(\mymat{\Omega}_{\iota}) \le n \sigma^+$ 
OR $\sigma^+(\mymat{\Omega}_{\pi}) \ge n \sigma^+$ (for any $n$ as the above).
Due to the inequality of Weyl (Theorem 1 in \citealt{stewart1998perturbation}; see below), 
the fact that $\mymat{\Omega}_{\pi} = \mymat{\Omega}_{\pi \mid K} + \mymat{\Omega}_{\pi \mid I(\pi)}$, Lemma~\ref{lemma:hoeff},
and the fact that $n-k \le n$, the probability of this OR event is bounded from above
by $4dd' \exp \left(- \displaystyle\frac{(n-k)(\sigma^+)^2}{(dd'B)^2} \right)$.


The conclusion from this is that if we were to sample uniformly a permutation $\pi$ from the set of permutations
over $[n]$, then with quite high likelihood (because the fraction of elements that are preserved under $\pi$ becomes smaller
as $n$ becomes larger), the sum of the singular values of $\mymat{\Omega}_{\pi}$ under this
permutation will be smaller than the sum of the singular values of $\mymat{\Omega}_{\iota}$ -- meaning, when the
$x$s and the $z$s are correctly aligned. This justifies our objective of aligning the $x$s and the $z$s with an
objective that maximizes the singular values, following Proposition~\ref{proposition:A}.

\paragraph{Inequality of Weyl (1912)}
As mentioned by \newcite{stewart1998perturbation}, the following holds:

\begin{lemma}\label{lem:weyl}
Let $\mymat{A}$ and $\mymat{E}$ be two matrices, and let $\mymat{\tilde{A}} = \mymat{A} + \mymat{E}$.
Let $\sigma_i$ be the $i$th singular value of $\mymat{A}$
and $\tilde{\sigma}_i$ be the $i$th singular value of $\mymat{\tilde{A}}$.
Then $|\sigma_i - \tilde{\sigma}_i| \le ||\mymat{E}||_2$.
\end{lemma}

\section{Comprehensive Results on the BiasBench Datasets}
\label{appendix:benchbias}

We include more results for the SEAT dataset from BiasBench and for the CrowS-Pairs
dataset and StereoSet datasets for bias categories other than gender. A description of
the SEAT and GLUE datasets (with metrics used) follows.

\paragraph{SEAT \cite{may-etal-2019-measuring}}
SEAT is a sentence-level extension of WEAT \cite{caliskan2017semantics},    which is an association test between two categories of words: attribute word sets and target word sets. For example, attribute words for gender bias could be \{ \emph{he}, \emph{man} \}, while a target words could be  \{ \emph{career}, \emph{office} \}. 
For example, an attribute word set (in case of gender bias) could be a set of words such
as \{ \emph{he}, \emph{him}, \emph{man} \}, while a target
word set might be words related to office work.
If we see a high association between an attribute word set and
a target word set, we may claim that a particular gender bias is encoded.
The final evaluation is calculated by measuring the similarity between the different attributes and target word sets. To extend WEAT to a sentence-level test, \cite{caliskan2017semantics} incorporated the WEAT attribute and target words into synthetic sentence templates. 

We use an \emph{effect size} metric to report our results for SEAT. This measure is a normalized difference between cosine similarity of representations of the attribute words and the target words. Both attribute words and target words are split into two categories (for example, in relation to gender), so the difference is based on four terms, between each pair of each category set of words (target and attribute). An effect size closer to zero indicates less bias is encoded in the representations.

\paragraph{GLUE \cite{wang2018glue}}
We follow \newcite{meade2022empirical} and use the GLUE dataset to test the debiased model on an array of downstream tasks to validate their usability. GLUE is a highly popular benchmark for testing NLP models, containing a variety of tasks, such as classification tasks (e.g., sentiment analysis), similarity tasks (e.g., paraphrase identification), and inference tasks (e.g., question-answering). 

The following tables of results are included:

\begin{itemize}
    \item Table~\ref{StereoSet_race_religion} presents the StereoSet results for removing the race (a) and religion (b) guarded attributes. 
    \item Tables~\ref{tab:SEAT-gender}, \ref{tab:SEAT-race}, and \ref{tab:SEAT-religion} describe the SEAT effect sizes for the gender, race, and religion cases, respectively. 
    \item Table~\ref{tab:glue} presents the scores the debiased representations achieve for the GLUE benchmark.
\end{itemize}

\begin{table*}[]

\begin{tabular}{cc}

\begin{minipage}{0.5\textwidth}
\scalebox{0.9}{

\centering
\begin{tabular}{lrr}
\toprule
                   Model & S. Score (\%) &  LM Score (\%) \\
\midrule
                                Race\\
\midrule
                                    BERT &                   57.03 &                       84.17 \\
        $$\,$$ + \textsc{AM} + \textsc{INLP} &         \ua{1.23} 58.26 &            \dab{0.65} 83.53 \\
    $$\,$$ + \textsc{Kmeans} + \textsc{INLP} &         \ua{0.32} 57.35 &            \dab{0.63} 83.54 \\
       $\,$ + \textsc{Oracle}\textsc{INLP} &         \ua{0.33} 57.36 &            \dab{1.05} 83.12 \\
      $\,$ + \textsc{Partial}\textsc{INLP} &         \ua{0.33} 57.36 &            \dab{1.05} 83.12 \\
    $\,$ + \textsc{AMSAL}&         \ua{1.98} 59.01 &            \uag{0.55} 84.72 \\
$\,$ + \textsc{Kmeans} + \textsc{SAL} &         \ua{2.10} 59.13 &            \uag{0.49} 84.66 \\
   $\,$ + \textsc{Oracle}\textsc{SAL} &         \ua{1.85} 58.88 &            \uag{0.76} 84.93 \\
  $\,$ + \textsc{Partial}\textsc{SAL} &         \ua{1.85} 58.88 &            \uag{0.76} 84.93 \\
  
\midrule
                                  ALBERT &                   57.57 &                       89.77 \\
        $\,$ + \textsc{AM} + \textsc{INLP} &         \da{0.94} 56.63 &            \dab{2.25} 87.52 \\
    $\,$ + \textsc{Kmeans} + \textsc{INLP} &         \ua{1.42} 59.00 &            \dab{2.04} 87.72 \\
       $\,$ + \textsc{Oracle}\textsc{INLP} &         \da{2.54} 55.04 &            \dab{1.95} 87.82 \\
      $\,$ + \textsc{Partial}\textsc{INLP} &         \da{2.54} 55.04 &            \dab{1.97} 87.80 \\
    $\,$ + \textsc{AMSAL}&         \da{0.84} 56.73 &            \uag{0.09} 89.86 \\
$\,$ + \textsc{Kmeans} + \textsc{SAL} &         \da{1.31} 56.27 &            \uag{0.06} 89.82 \\
   $\,$ + \textsc{Oracle}\textsc{SAL} &         \da{0.31} 57.26 &            \uag{0.75} 90.52 \\
  $\,$ + \textsc{Partial}\textsc{SAL} &         \da{0.31} 57.26 &            \uag{0.75} 90.52 \\
  
\midrule
                                   GPT-2 &                   58.83 &                       91.01 \\
        $\,$ + \textsc{AM} + \textsc{INLP} &         \da{0.90} 57.93 &            \dab{5.55} 85.47 \\
    $\,$ + \textsc{Kmeans} + \textsc{INLP} &         \ua{0.10} 58.93 &            \dab{0.94} 90.07 \\
       $\,$ + \textsc{Oracle}\textsc{INLP} &         \ua{0.21} 59.04 &            \uag{0.04} 91.06 \\
      $\,$ + \textsc{Partial}\textsc{INLP} &         \ua{0.21} 59.04 &            \uag{0.03} 91.05 \\
    $\,$ + \textsc{AMSAL}&         \da{3.15} 55.69 &            \dab{0.43} 90.59 \\
$\,$ + \textsc{Kmeans} + \textsc{SAL} &         \da{2.65} 56.18 &            \uag{0.08} 91.09 \\
   $\,$ + \textsc{Oracle}\textsc{SAL} &         \da{3.09} 55.75 &            \dab{2.09} 88.92 \\
  $\,$ + \textsc{Partial}\textsc{SAL} &         \da{3.09} 55.75 &            \dab{2.09} 88.92 \\
  
\midrule
                                 RoBERTa &                   61.67 &                       88.95 \\
        $\,$ + \textsc{AM} + \textsc{INLP} &         \da{7.31} 54.37 &            \dab{3.06} 85.88 \\
    $\,$ + \textsc{Kmeans} + \textsc{INLP} &         \da{5.67} 56.00 &            \dab{3.95} 85.00 \\
       $\,$ + \textsc{Oracle}\textsc{INLP} &         \da{3.42} 58.26 &            \uag{0.01} 88.96 \\
      $\,$ + \textsc{Partial}\textsc{INLP} &         \da{3.42} 58.26 &            \uag{0.02} 88.96 \\
    $\,$ + \textsc{AMSAL}&         \ua{0.74} 62.41 &            \uag{1.02} 89.96 \\
$\,$ + \textsc{Kmeans} + \textsc{SAL} &         \ua{0.71} 62.39 &            \uag{1.03} 89.97 \\
   $\,$ + \textsc{Oracle}\textsc{SAL} &         \ua{1.79} 63.47 &            \uag{0.49} 89.44 \\
  $\,$ + \textsc{Partial}\textsc{SAL} &         \ua{1.79} 63.47 &            \uag{0.49} 89.44 \\

\bottomrule
\end{tabular}
}

\end{minipage}
&

\begin{minipage}{0.5\textwidth}
\scalebox{0.9}{
\begin{tabular}{lrr}
\toprule
                   Model & S. Score (\%) &  LM Score (\%) \\
\midrule
                                Religion\\
\midrule
                                    BERT &                   59.70 &                       84.17 \\
        $\,$ + \textsc{AM} + \textsc{INLP} &         \ua{3.22} 62.92 &            \dab{0.60} 83.58 \\
    $\,$ + \textsc{Kmeans} + \textsc{INLP} &         \ua{1.67} 61.38 &            \dab{0.28} 83.89 \\
       $\,$ + \textsc{Oracle}\textsc{INLP} &         \ua{0.61} 60.31 &            \dab{0.82} 83.35 \\
      $\,$ + \textsc{Partial}\textsc{INLP} &         \ua{0.61} 60.31 &            \dab{0.82} 83.35 \\
    $\,$ + \textsc{AMSAL}&         \ua{1.42} 61.12 &            \uag{0.60} 84.77 \\
$\,$ + \textsc{Kmeans} + \textsc{SAL} &         \ua{1.83} 61.53 &            \uag{0.62} 84.79 \\
   $\,$ + \textsc{Oracle}\textsc{SAL} &         \ua{0.09} 59.79 &            \uag{0.68} 84.85 \\
  $\,$ + \textsc{Partial}\textsc{SAL} &         \ua{0.09} 59.79 &            \uag{0.68} 84.85 \\
  
\midrule
                                  ALBERT &                   60.32 &                       89.77 \\
        $\,$ + \textsc{AM} + \textsc{INLP} &         \ua{1.85} 62.17 &            \dab{1.20} 88.57 \\
    $\,$ + \textsc{Kmeans} + \textsc{INLP} &         \ua{2.78} 63.10 &            \dab{1.41} 88.36 \\
       $\,$ + \textsc{Oracle}\textsc{INLP} &         \ua{3.45} 63.77 &            \dab{0.91} 88.86 \\
      $\,$ + \textsc{Partial}\textsc{INLP} &         \ua{3.45} 63.77 &            \dab{0.91} 88.86 \\
    $\,$ + \textsc{AMSAL}&         \da{0.78} 59.54 &            \uag{0.39} 90.15 \\
$\,$ + \textsc{Kmeans} + \textsc{SAL} &         \da{0.39} 59.94 &            \uag{0.36} 90.13 \\
   $\,$ + \textsc{Oracle}\textsc{SAL} &         \da{1.18} 59.14 &            \uag{1.26} 91.02 \\
  $\,$ + \textsc{Partial}\textsc{SAL} &         \da{1.18} 59.14 &            \uag{1.26} 91.02 \\
  
\midrule
                                   GPT-2 &                   63.26 &                       91.01 \\
        $\,$ + \textsc{AM} + \textsc{INLP} &         \da{1.90} 61.36 &            \dab{6.57} 84.44 \\
    $\,$ + \textsc{Kmeans} + \textsc{INLP} &         \ua{1.66} 64.92 &            \dab{0.88} 90.14 \\
       $\,$ + \textsc{Oracle}\textsc{INLP} &         \ua{0.69} 63.95 &            \uag{0.19} 91.21 \\
      $\,$ + \textsc{Partial}\textsc{INLP} &         \ua{0.69} 63.95 &            \uag{0.19} 91.21 \\
    $\,$ + \textsc{AMSAL}&         \da{4.83} 58.43 &            \dab{0.60} 90.41 \\
$\,$ + \textsc{Kmeans} + \textsc{SAL} &         \da{2.57} 60.69 &            \dab{0.41} 90.60 \\
   $\,$ + \textsc{Oracle}\textsc{SAL} &         \da{5.48} 57.78 &            \dab{3.91} 87.10 \\
  $\,$ + \textsc{Partial}\textsc{SAL} &         \da{5.48} 57.78 &            \dab{3.91} 87.10 \\
\midrule
                                 RoBERTa &                   64.28 &                       88.95 \\
        $\,$ + \textsc{AM} + \textsc{INLP} &         \da{3.84} 60.44 &            \dab{4.09} 84.86 \\
    $\,$ + \textsc{Kmeans} + \textsc{INLP} &         \da{1.37} 62.91 &            \dab{2.82} 86.13 \\
       $\,$ + \textsc{Oracle}\textsc{INLP} &         \da{3.94} 60.34 &            \dab{0.83} 88.12 \\
      $\,$ + \textsc{Partial}\textsc{INLP} &         \da{3.94} 60.34 &            \dab{0.84} 88.11 \\
    $\,$ + \textsc{AMSAL}&         \da{1.64} 62.64 &            \uag{1.00} 89.95 \\
$\,$ + \textsc{Kmeans} + \textsc{SAL} &         \da{2.04} 62.24 &            \uag{0.99} 89.93 \\
   $\,$ + \textsc{Oracle}\textsc{SAL} &         \da{1.92} 62.36 &            \uag{1.11} 90.06 \\
  $\,$ + \textsc{Partial}\textsc{SAL} &         \da{1.92} 62.36 &            \uag{1.11} 90.06 \\

\bottomrule
\end{tabular}
}
\end{minipage}

\\ 

(a) & (b)

\end{tabular}

\caption{(a) StereoSet stereotype scores and language modeling scores (LM Score) for race debiased BERT, ALBERT, RoBERTa, and GPT-2 models. Stereotype scores are least biased at 50\% and the LM Scores are best at 100\%; (b) StereoSet stereotype scores and language modeling scores (LM Score) for religion debiased BERT, ALBERT, RoBERTa, and GPT-2 models. Stereotype scores are least biased at 50\% and the LM Scores are best at 100\%.}
\label{StereoSet_race_religion}
\end{table*}

\clearpage

\begin{table*}[]
\centering
\footnotesize
\begin{tabular}{lrrrrrrr}
\toprule
                   Model &   SEAT6 & SEAT6b & SEAT7 &  SEAT7b &   SEAT8 &  SEAT8b & Avg. Effect Size \\
\midrule
                                    BERT & 0.931 {$^*$} &       0.090 &     -0.124 & 0.937 {$^*$} & 0.783 {$^*$} & 0.858 {$^*$} &                           0.620 \\
        $\,$ + \textsc{AM} + \textsc{INLP} & 0.744 {$^*$} &      -0.006 &      0.036 & 0.968 {$^*$} & 0.828 {$^*$} & 0.849 {$^*$} &                \da{0.049} 0.572 \\
    $\,$ + \textsc{Kmeans} + \textsc{INLP} & 0.809 {$^*$} &       0.013 &     -0.084 & 0.812 {$^*$} & 0.756 {$^*$} & 0.785 {$^*$} &                \da{0.077} 0.543 \\
       $\,$ + \textsc{Oracle}\textsc{INLP} &        0.269 &      -0.339 &     -0.403 & 0.437 {$^*$} &        0.399 &        0.289 &                \da{0.264} 0.356 \\
      $\,$ + \textsc{Partial}\textsc{INLP} &        0.269 &      -0.338 &     -0.404 & 0.436 {$^*$} &        0.399 &        0.289 &                \da{0.265} 0.356 \\
    $\,$ + \textsc{AMSAL}& 0.928 {$^*$} &       0.110 &     -0.191 & 0.717 {$^*$} & 0.756 {$^*$} & 0.756 {$^*$} &                \da{0.044} 0.576 \\
$\,$ + \textsc{Kmeans} + \textsc{SAL} & 0.925 {$^*$} &       0.109 &     -0.190 & 0.722 {$^*$} & 0.752 {$^*$} & 0.752 {$^*$} &                \da{0.046} 0.575 \\
   $\,$ + \textsc{Oracle}\textsc{SAL} &        0.387 &      -0.301 &     -0.876 &       -0.192 &        0.299 &        0.309 &                \da{0.227} 0.394 \\
  $\,$ + \textsc{Partial}\textsc{SAL} &        0.387 &      -0.301 &     -0.876 &       -0.192 &        0.299 &        0.309 &                \da{0.227} 0.394 \\

\midrule
                                  ALBERT & 0.637 {$^*$} &       0.151 & 0.487 {$^*$} & 0.956 {$^*$} & 0.683 {$^*$} & 0.823 {$^*$} &                           0.623 \\
        $\,$ + \textsc{AM} + \textsc{INLP} & 0.620 {$^*$} &       0.165 & 0.408 {$^*$} & 0.854 {$^*$} & 0.649 {$^*$} & 0.744 {$^*$} &                \da{0.049} 0.573 \\
    $\,$ + \textsc{Kmeans} + \textsc{INLP} & 0.645 {$^*$} &       0.147 & 0.408 {$^*$} & 0.822 {$^*$} & 0.660 {$^*$} & 0.829 {$^*$} &                \da{0.038} 0.585 \\
       $\,$ + \textsc{Oracle}\textsc{INLP} & 0.464 {$^*$} &      -0.084 &       -0.222 & 0.467 {$^*$} &        0.215 & 0.462 {$^*$} &                \da{0.304} 0.319 \\
      $\,$ + \textsc{Partial}\textsc{INLP} & 0.464 {$^*$} &      -0.084 &       -0.222 & 0.467 {$^*$} &        0.215 & 0.462 {$^*$} &                \da{0.304} 0.319 \\
    $\,$ + \textsc{AMSAL}& 0.640 {$^*$} &       0.138 & 0.477 {$^*$} & 0.933 {$^*$} & 0.666 {$^*$} & 0.820 {$^*$} &                \da{0.010} 0.612 \\
$\,$ + \textsc{Kmeans} + \textsc{SAL} & 0.640 {$^*$} &       0.136 & 0.474 {$^*$} & 0.933 {$^*$} & 0.664 {$^*$} & 0.818 {$^*$} &                \da{0.012} 0.611 \\
   $\,$ + \textsc{Oracle}\textsc{SAL} & 0.468 {$^*$} &      -0.067 &       -0.230 &        0.312 &        0.305 & 0.545 {$^*$} &                \da{0.302} 0.321 \\
  $\,$ + \textsc{Partial}\textsc{SAL} & 0.468 {$^*$} &      -0.067 &       -0.230 &        0.312 &        0.305 & 0.545 {$^*$} &                \da{0.302} 0.321 \\

\midrule
                                 RoBERTa & 0.922 {$^*$} &       0.208 & 0.979 {$^*$} & 1.460 {$^*$} & 0.810 {$^*$} & 1.261 {$^*$} &                           0.940 \\
        $\,$ + \textsc{AM} + \textsc{INLP} & 0.982 {$^*$} &       0.262 & 0.845 {$^*$} & 1.575 {$^*$} & 0.840 {$^*$} & 1.395 {$^*$} &                \ua{0.043} 0.983 \\
    $\,$ + \textsc{Kmeans} + \textsc{INLP} & 0.933 {$^*$} &       0.238 & 1.090 {$^*$} & 1.595 {$^*$} & 1.148 {$^*$} & 1.435 {$^*$} &                \ua{0.133} 1.073 \\
       $\,$ + \textsc{Oracle}\textsc{INLP} & 0.781 {$^*$} &       0.014 & 0.651 {$^*$} & 1.281 {$^*$} & 0.708 {$^*$} & 1.160 {$^*$} &                \da{0.174} 0.766 \\
      $\,$ + \textsc{Partial}\textsc{INLP} & 0.782 {$^*$} &       0.014 & 0.651 {$^*$} & 1.282 {$^*$} & 0.708 {$^*$} & 1.161 {$^*$} &                \da{0.174} 0.766 \\
    $\,$ + \textsc{AMSAL}& 0.902 {$^*$} &       0.187 & 1.021 {$^*$} & 1.549 {$^*$} & 0.893 {$^*$} & 1.386 {$^*$} &                \ua{0.050} 0.990 \\
$\,$ + \textsc{Kmeans} + \textsc{SAL} & 0.920 {$^*$} &       0.182 & 1.017 {$^*$} & 1.549 {$^*$} & 0.885 {$^*$} & 1.389 {$^*$} &                \ua{0.051} 0.990 \\
   $\,$ + \textsc{Oracle}\textsc{SAL} & 0.695 {$^*$} &      -0.014 & 0.550 {$^*$} & 1.315 {$^*$} & 0.684 {$^*$} & 1.170 {$^*$} &                \da{0.202} 0.738 \\
  $\,$ + \textsc{Partial}\textsc{SAL} & 0.695 {$^*$} &      -0.014 & 0.550 {$^*$} & 1.315 {$^*$} & 0.684 {$^*$} & 1.170 {$^*$} &                \da{0.202} 0.738 \\

\midrule
                                   GPT-2 &      0.138 &       0.003 &       -0.023 &       0.002 &     -0.224 &      -0.287 &                           0.113 \\
        $\,$ + \textsc{AM} + \textsc{INLP} &      0.141 &       0.009 &       -0.017 &       0.010 &     -0.213 &      -0.283 &                \da{0.001} 0.112 \\
    $\,$ + \textsc{Kmeans} + \textsc{INLP} &      0.138 &       0.006 &       -0.024 &       0.002 &     -0.223 &      -0.287 &                \ua{0.000} 0.113 \\
       $\,$ + \textsc{Oracle}\textsc{INLP} &      0.130 &      -0.005 &       -0.024 &       0.000 &     -0.229 &      -0.291 &                \ua{0.000} 0.113 \\
      $\,$ + \textsc{Partial}\textsc{INLP} &      0.130 &      -0.005 &       -0.024 &       0.000 &     -0.229 &      -0.291 &                \ua{0.000} 0.113 \\
    $\,$ + \textsc{AMSAL}&      0.280 &       0.199 & 0.900 {$^*$} &       0.352 &      0.408 &       0.118 &                \ua{0.263} 0.376 \\
$\,$ + \textsc{Kmeans} + \textsc{SAL} &      0.298 &      -0.304 &        0.030 &       0.044 &      0.114 &      -0.096 &                \ua{0.035} 0.148 \\
   $\,$ + \textsc{Oracle}\textsc{SAL} &      0.155 &      -0.379 &       -0.093 &      -0.059 &     -0.039 &      -0.173 &                \ua{0.037} 0.150 \\
  $\,$ + \textsc{Partial}\textsc{SAL} &      0.155 &      -0.379 &       -0.093 &      -0.059 &     -0.039 &      -0.173 &                \ua{0.037} 0.150 \\

\bottomrule
\end{tabular}
\caption{SEAT effect sizes for gender-debiased representations of BERT, ALBERT, RoBERTa, and GPT-2 models. Effect sizes closer to 0 are indicative of less biased model representations. Statistically significant effect sizes at $p < 0.01$ are denoted by *. The final column reports the average absolute effect size across all six gender SEAT tests for each debiased model.}
\label{tab:SEAT-gender}
\end{table*}

\clearpage

\begin{table*}[]
\centering
\footnotesize
\begin{tabular}{lrrrrrrrr}
\toprule
                   Model & ABW-1 & ABW-2 &   SEAT-3 &  SEAT-3b &   SEAT-4 &   SEAT-5 &  SEAT-5b & Avg. Effect Size \\
\midrule
                BERT &                            -0.079 &                        0.690 {$^*$} & 0.778 {$^*$} & 0.469 {$^*$} & 0.901 {$^*$} & 0.887 {$^*$} & 0.539 {$^*$} &                           0.620 \\
        $\,$ + \textsc{AM} + \textsc{INLP} &                             0.155 &                        0.583 {$^*$} & 0.769 {$^*$} & 0.341 {$^*$} & 0.889 {$^*$} & 0.937 {$^*$} & 0.403 {$^*$} &                \da{0.038} 0.582 \\
    $\,$ + \textsc{Kmeans} + \textsc{INLP} &                             0.097 &                        0.590 {$^*$} & 0.775 {$^*$} & 0.381 {$^*$} & 0.882 {$^*$} & 0.888 {$^*$} & 0.357 {$^*$} &                \da{0.053} 0.567 \\
       $\,$ + \textsc{Oracle}\textsc{INLP} &                             0.295 &                        0.565 {$^*$} & 0.799 {$^*$} & 0.369 {$^*$} & 0.977 {$^*$} & 1.039 {$^*$} & 0.432 {$^*$} &                \ua{0.019} 0.639 \\
      $\,$ + \textsc{Partial}\textsc{INLP} &                             0.295 &                        0.565 {$^*$} & 0.799 {$^*$} & 0.370 {$^*$} & 0.976 {$^*$} & 1.039 {$^*$} & 0.432 {$^*$} &                \ua{0.019} 0.639 \\
    $\,$ + \textsc{AMSAL}&                             0.138 &                        0.621 {$^*$} & 0.797 {$^*$} & 0.374 {$^*$} & 0.911 {$^*$} & 1.015 {$^*$} & 0.435 {$^*$} &                \da{0.007} 0.613 \\
$\,$ + \textsc{Kmeans} + \textsc{SAL} &                             0.131 &                        0.624 {$^*$} & 0.798 {$^*$} & 0.373 {$^*$} & 0.912 {$^*$} & 1.013 {$^*$} & 0.436 {$^*$} &                \da{0.008} 0.612 \\
   $\,$ + \textsc{Oracle}\textsc{SAL} &                            -0.021 &                        0.643 {$^*$} & 0.788 {$^*$} & 0.357 {$^*$} & 0.885 {$^*$} & 0.894 {$^*$} & 0.431 {$^*$} &                \da{0.046} 0.574 \\
  $\,$ + \textsc{Partial}\textsc{SAL} &                            -0.021 &                        0.643 {$^*$} & 0.788 {$^*$} & 0.357 {$^*$} & 0.885 {$^*$} & 0.894 {$^*$} & 0.431 {$^*$} &                \da{0.046} 0.574 \\

\midrule
              ALBERT &                            -0.014 &                               0.410 & 1.132 {$^*$} &      -0.252 & 0.956 {$^*$} & 1.041 {$^*$} &       0.058 &                           0.552 \\
        $\,$ + \textsc{AM} + \textsc{INLP} &                            -0.150 &                        0.505 {$^*$} & 1.149 {$^*$} &      -0.244 & 0.982 {$^*$} & 1.075 {$^*$} &      -0.036 &                \ua{0.040} 0.592 \\
    $\,$ + \textsc{Kmeans} + \textsc{INLP} &                            -0.015 &                        0.484 {$^*$} & 1.162 {$^*$} &      -0.228 & 0.988 {$^*$} & 1.067 {$^*$} &      -0.033 &                \ua{0.017} 0.568 \\
       $\,$ + \textsc{Oracle}\textsc{INLP} &                             0.040 &                        0.534 {$^*$} & 1.165 {$^*$} &      -0.150 & 0.996 {$^*$} & 1.116 {$^*$} &       0.021 &                \ua{0.023} 0.574 \\
      $\,$ + \textsc{Partial}\textsc{INLP} &                             0.040 &                        0.534 {$^*$} & 1.165 {$^*$} &      -0.150 & 0.996 {$^*$} & 1.116 {$^*$} &       0.021 &                \ua{0.023} 0.574 \\
    $\,$ + \textsc{AMSAL}&                             0.283 &                        0.471 {$^*$} & 0.985 {$^*$} &      -0.299 & 0.802 {$^*$} & 0.938 {$^*$} &      -0.063 &                \da{0.003} 0.549 \\
$\,$ + \textsc{Kmeans} + \textsc{SAL} &                             0.292 &                        0.472 {$^*$} & 0.980 {$^*$} &      -0.294 & 0.799 {$^*$} & 0.935 {$^*$} &      -0.060 &                \da{0.004} 0.547 \\
   $\,$ + \textsc{Oracle}\textsc{SAL} &                      0.300 {$^*$} &                        0.471 {$^*$} & 0.994 {$^*$} &      -0.281 & 0.813 {$^*$} & 0.949 {$^*$} &      -0.089 &                \ua{0.005} 0.557 \\
  $\,$ + \textsc{Partial}\textsc{SAL} &                      0.300 {$^*$} &                        0.471 {$^*$} & 0.994 {$^*$} &      -0.281 & 0.813 {$^*$} & 0.949 {$^*$} &      -0.089 &                \ua{0.005} 0.557 \\

\midrule
            RoBERTa &                      0.395 {$^*$} &                               0.159 &       -0.114 &      -0.003 &       -0.315 & 0.780 {$^*$} & 0.386 {$^*$} &                           0.307 \\
        $\,$ + \textsc{AM} + \textsc{INLP} &                             0.257 &                        0.534 {$^*$} & 0.381 {$^*$} &       0.138 &        0.202 & 0.646 {$^*$} & 0.300 {$^*$} &                \ua{0.044} 0.351 \\
    $\,$ + \textsc{Kmeans} + \textsc{INLP} &                             0.270 &                        0.466 {$^*$} & 0.242 {$^*$} &       0.116 &        0.079 & 0.627 {$^*$} & 0.310 {$^*$} &                \da{0.006} 0.301 \\
       $\,$ + \textsc{Oracle}\textsc{INLP} &                             0.222 &                               0.445 & 0.354 {$^*$} &       0.130 &        0.125 & 0.636 {$^*$} & 0.301 {$^*$} &                \ua{0.009} 0.316 \\
      $\,$ + \textsc{Partial}\textsc{INLP} &                             0.222 &                               0.445 & 0.354 {$^*$} &       0.130 &        0.125 & 0.636 {$^*$} & 0.301 {$^*$} &                \ua{0.009} 0.316 \\
    $\,$ + \textsc{AMSAL}&                      0.317 {$^*$} &                        0.520 {$^*$} & 0.471 {$^*$} &       0.211 & 0.314 {$^*$} & 0.576 {$^*$} & 0.275 {$^*$} &                \ua{0.076} 0.384 \\
$\,$ + \textsc{Kmeans} + \textsc{SAL} &                      0.314 {$^*$} &                        0.522 {$^*$} & 0.476 {$^*$} &       0.212 & 0.320 {$^*$} & 0.579 {$^*$} & 0.276 {$^*$} &                \ua{0.078} 0.385 \\
   $\,$ + \textsc{Oracle}\textsc{SAL} &                             0.170 &                               0.220 & 0.401 {$^*$} &       0.227 &        0.225 & 0.471 {$^*$} & 0.268 {$^*$} &                \da{0.024} 0.283 \\
  $\,$ + \textsc{Partial}\textsc{SAL} &                             0.170 &                               0.220 & 0.401 {$^*$} &       0.227 &        0.225 & 0.471 {$^*$} & 0.268 {$^*$} &                \da{0.024} 0.283 \\

\midrule
             GPT-2 &                      1.060 {$^*$} &                              -0.200 & 0.431 {$^*$} & 0.243 {$^*$} &        0.133 & 0.696 {$^*$} & 0.370 {$^*$} &                           0.448 \\
        $\,$ + \textsc{AM} + \textsc{INLP} &                      1.046 {$^*$} &                              -0.169 & 0.472 {$^*$} & 0.257 {$^*$} &        0.172 & 0.686 {$^*$} & 0.366 {$^*$} &                \ua{0.005} 0.452 \\
    $\,$ + \textsc{Kmeans} + \textsc{INLP} &                      1.059 {$^*$} &                              -0.189 & 0.440 {$^*$} & 0.248 {$^*$} &        0.141 & 0.695 {$^*$} & 0.373 {$^*$} &                \ua{0.002} 0.449 \\
       $\,$ + \textsc{Oracle}\textsc{INLP} &                      1.060 {$^*$} &                              -0.200 & 0.433 {$^*$} & 0.246 {$^*$} &        0.135 & 0.693 {$^*$} & 0.364 {$^*$} &                \da{0.000} 0.447 \\
      $\,$ + \textsc{Partial}\textsc{INLP} &                      1.060 {$^*$} &                              -0.200 & 0.433 {$^*$} & 0.246 {$^*$} &        0.135 & 0.693 {$^*$} & 0.364 {$^*$} &                \da{0.000} 0.447 \\
    $\,$ + \textsc{AMSAL}&                            -0.525 &                               0.157 & 0.654 {$^*$} & 0.398 {$^*$} & 0.327 {$^*$} &        0.120 &        0.117 &                \da{0.119} 0.328 \\
$\,$ + \textsc{Kmeans} + \textsc{SAL} &                            -0.459 &                               0.179 & 1.088 {$^*$} & 0.499 {$^*$} & 0.827 {$^*$} & 0.594 {$^*$} & 0.271 {$^*$} &                \ua{0.112} 0.560 \\
   $\,$ + \textsc{Oracle}\textsc{SAL} &                             0.183 &                               0.101 & 1.095 {$^*$} & 0.515 {$^*$} & 0.836 {$^*$} & 0.817 {$^*$} & 0.347 {$^*$} &                \ua{0.109} 0.556 \\
  $\,$ + \textsc{Partial}\textsc{SAL} &                             0.183 &                               0.101 & 1.095 {$^*$} & 0.515 {$^*$} & 0.836 {$^*$} & 0.817 {$^*$} & 0.347 {$^*$} &                \ua{0.109} 0.556 \\

\bottomrule
\end{tabular}
\caption{SEAT effect sizes for race debiased BERT, ALBERT, RoBERTa, and GPT-2 models. Effect sizes closer to 0 are indicative of less biased model representations. Statistically significant effect sizes at $p < 0.01$ are denoted by *. The final column reports the average absolute effect size across all six gender SEAT tests for each debiased model.}
\label{tab:SEAT-race}
\end{table*}

\clearpage

\begin{table*}[]
\centering
\begin{tabular}{lrrrrrrr}
\toprule
                   Model & Religion-1 & Religion-1b & Religion-2 & Religion-2b & Avg. Effect Size \\
\midrule
                BERT &   0.744 {$^*$} &          -0.067 &   1.009 {$^*$} &          -0.147 &                           0.492 \\
        $\,$ + \textsc{AM} + \textsc{INLP} &   0.530 {$^*$} &          -0.184 &   0.847 {$^*$} &          -0.156 &                \da{0.062} 0.429 \\
    $\,$ + \textsc{Kmeans} + \textsc{INLP} &   0.574 {$^*$} &          -0.253 &   0.919 {$^*$} &          -0.308 &                \ua{0.022} 0.514 \\
       $\,$ + \textsc{Oracle}\textsc{INLP} &   0.473 {$^*$} &          -0.301 &   0.787 {$^*$} &          -0.280 &                \da{0.031} 0.460 \\
      $\,$ + \textsc{Partial}\textsc{INLP} &   0.473 {$^*$} &          -0.301 &   0.787 {$^*$} &          -0.280 &                \da{0.031} 0.460 \\
    $\,$ + \textsc{AMSAL}&   0.683 {$^*$} &          -0.117 &   0.941 {$^*$} &          -0.178 &                \da{0.012} 0.480 \\
$\,$ + \textsc{Kmeans} + \textsc{SAL} &   0.686 {$^*$} &          -0.112 &   0.938 {$^*$} &          -0.180 &                \da{0.013} 0.479 \\
   $\,$ + \textsc{Oracle}\textsc{SAL} &   0.735 {$^*$} &          -0.036 &   0.884 {$^*$} &          -0.156 &                \da{0.039} 0.453 \\
  $\,$ + \textsc{Partial}\textsc{SAL} &   0.735 {$^*$} &          -0.036 &   0.884 {$^*$} &          -0.156 &                \da{0.039} 0.453 \\

\midrule
          ALBERT &          0.203 &          -0.117 &   0.848 {$^*$} &    0.555 {$^*$} &                           0.431 \\
        $\,$ + \textsc{AM} + \textsc{INLP} &          0.208 &          -0.065 &   0.891 {$^*$} &    0.557 {$^*$} &                \da{0.001} 0.430 \\
    $\,$ + \textsc{Kmeans} + \textsc{INLP} &          0.126 &          -0.138 &   0.839 {$^*$} &    0.518 {$^*$} &                \da{0.025} 0.405 \\
       $\,$ + \textsc{Oracle}\textsc{INLP} &          0.206 &          -0.110 &   0.727 {$^*$} &    0.385 {$^*$} &                \da{0.074} 0.357 \\
      $\,$ + \textsc{Partial}\textsc{INLP} &          0.206 &          -0.110 &   0.727 {$^*$} &    0.385 {$^*$} &                \da{0.074} 0.357 \\
    $\,$ + \textsc{AMSAL}&          0.024 &          -0.256 &   0.722 {$^*$} &    0.418 {$^*$} &                \da{0.076} 0.355 \\
$\,$ + \textsc{Kmeans} + \textsc{SAL} &          0.027 &          -0.253 &   0.722 {$^*$} &    0.415 {$^*$} &                \da{0.077} 0.354 \\
   $\,$ + \textsc{Oracle}\textsc{SAL} &          0.116 &          -0.168 &   0.585 {$^*$} &           0.289 &                \da{0.141} 0.289 \\
  $\,$ + \textsc{Partial}\textsc{SAL} &          0.116 &          -0.168 &   0.585 {$^*$} &           0.289 &                \da{0.141} 0.289 \\

\midrule
        RoBERTa &          0.132 &           0.018 &         -0.191 &          -0.166 &                           0.127 \\
        $\,$ + \textsc{AM} + \textsc{INLP} &         -0.042 &          -0.203 &         -0.255 &          -0.273 &                \ua{0.067} 0.193 \\
    $\,$ + \textsc{Kmeans} + \textsc{INLP} &         -0.014 &          -0.204 &         -0.187 &          -0.304 &                \ua{0.051} 0.177 \\
       $\,$ + \textsc{Oracle}\textsc{INLP} &         -0.309 &          -0.347 &         -0.191 &          -0.135 &                \ua{0.119} 0.246 \\
      $\,$ + \textsc{Partial}\textsc{INLP} &         -0.309 &          -0.347 &         -0.191 &          -0.135 &                \ua{0.119} 0.246 \\
    $\,$ + \textsc{AMSAL}&         -0.169 &          -0.228 &         -0.014 &           0.009 &                \da{0.022} 0.105 \\
$\,$ + \textsc{Kmeans} + \textsc{SAL} &         -0.172 &          -0.231 &         -0.011 &           0.011 &                \da{0.020} 0.106 \\
   $\,$ + \textsc{Oracle}\textsc{SAL} &         -0.063 &          -0.208 &         -0.203 &          -0.109 &                \ua{0.019} 0.146 \\
  $\,$ + \textsc{Partial}\textsc{SAL} &         -0.063 &          -0.208 &         -0.203 &          -0.109 &                \ua{0.019} 0.146 \\

\midrule
        GPT-2 &         -0.332 &          -0.271 &   0.617 {$^*$} &           0.286 &                           0.376 \\
        $\,$ + \textsc{AM} + \textsc{INLP} &         -0.326 &          -0.264 &   0.671 {$^*$} &           0.333 &                \ua{0.022} 0.399 \\
    $\,$ + \textsc{Kmeans} + \textsc{INLP} &         -0.331 &          -0.271 &   0.624 {$^*$} &           0.297 &                \ua{0.004} 0.380 \\
       $\,$ + \textsc{Oracle}\textsc{INLP} &         -0.331 &          -0.271 &   0.615 {$^*$} &           0.284 &                \da{0.001} 0.375 \\
      $\,$ + \textsc{Partial}\textsc{INLP} &         -0.331 &          -0.271 &   0.615 {$^*$} &           0.284 &                \da{0.001} 0.375 \\
    $\,$ + \textsc{AMSAL}&          0.087 &           0.064 &   0.767 {$^*$} &           0.341 &                \da{0.062} 0.315 \\
$\,$ + \textsc{Kmeans} + \textsc{SAL} &          0.274 &           0.300 &   1.014 {$^*$} &    0.534 {$^*$} &                \ua{0.154} 0.531 \\
   $\,$ + \textsc{Oracle}\textsc{SAL} &         -0.101 &          -0.089 &   1.144 {$^*$} &    0.779 {$^*$} &                \ua{0.152} 0.528 \\
  $\,$ + \textsc{Partial}\textsc{SAL} &         -0.101 &          -0.089 &   1.144 {$^*$} &    0.779 {$^*$} &                \ua{0.152} 0.528 \\

\bottomrule
\end{tabular}
\caption{SEAT effect sizes for religion debiased BERT, ALBERT, RoBERTa, and GPT-2 models. Effect sizes closer to 0 are indicative of less biased model representations. Statistically significant effect sizes at $p < 0.01$ are denoted by *. The final column reports the average absolute effect size across all six gender SEAT tests for each debiased model.}
\label{tab:SEAT-religion}
\end{table*}

\clearpage

\begin{table*}[]
\centering
\scalebox{0.9}{
\begin{tabular}{lrrrrrrrrrl}
\toprule
    Model &   cola &   mnli &   mrpc &   qnli &    qqp &    rte &   sst2 &   stsb &   wnli & Average \\
\midrule

                                BERT & 56.50 & 84.73 & 87.67 & 91.35 & 91.00 & 64.38 & 92.55 & 88.51 & 44.60 &               77.92 \\
    $\,$ + \textsc{AM} + \textsc{INLP} & 57.69 & 84.67 & 88.75 & 91.24 & 90.88 & 64.38 & 92.74 & 88.80 & 42.25 &    \uag{0.01} 77.93 \\
$\,$ + \textsc{Kmeans} + \textsc{INLP} & 56.84 & 84.72 & 88.23 & 91.35 & 90.94 & 64.02 & 92.62 & 88.61 & 40.38 &    \dab{0.40} 77.52 \\
   $\,$ + \textsc{Oracle}\textsc{INLP} & 57.44 & 84.62 & 88.32 & 91.37 & 91.02 & 63.66 & 92.74 & 88.32 & 46.48 &    \uag{0.30} 78.22 \\
  $\,$ + \textsc{Partial}\textsc{INLP} & 57.36 & 84.68 & 88.09 & 91.36 & 91.05 & 65.46 & 92.47 & 88.74 & 43.19 &    \uag{0.12} 78.04 \\
        $\,$ + \textsc{AM}\textsc{SAL} & 57.19 & 84.85 & 88.48 & 91.28 & 90.96 & 63.78 & 92.66 & 88.75 & 37.56 &    \dab{0.64} 77.28 \\
 $\,$ + \textsc{Kmeans} + \textsc{SAL} & 56.94 & 84.76 & 88.68 & 91.43 & 90.91 & 64.02 & 92.66 & 88.74 & 37.56 &    \dab{0.62} 77.30 \\
    $\,$ + \textsc{Oracle}\textsc{SAL} & 56.68 & 84.72 & 88.18 & 91.24 & 90.93 & 64.50 & 92.74 & 88.67 & 41.78 &    \dab{0.20} 77.72 \\
   $\,$ + \textsc{Partial}\textsc{SAL} & 56.16 & 84.71 & 87.82 & 91.31 & 90.92 & 64.50 & 92.58 & 88.70 & 41.78 &    \dab{0.31} 77.61 \\

\midrule

                              ALBERT & 46.55 & 85.31 & 91.17 & 91.73 & 90.82 & 70.88 & 91.55 & 90.57 & 43.66 &               78.03 \\
    $\,$ + \textsc{AM} + \textsc{INLP} & 57.30 & 85.46 & 90.57 & 91.63 & 90.49 & 70.28 & 91.86 & 90.62 & 48.83 &    \uag{1.65} 79.67 \\
$\,$ + \textsc{Kmeans} + \textsc{INLP} & 55.91 & 85.59 & 90.74 & 91.70 & 90.62 & 68.71 & 91.86 & 90.94 & 46.48 &    \uag{1.15} 79.17 \\
   $\,$ + \textsc{Oracle}\textsc{INLP} & 55.43 & 85.28 & 91.27 & 91.65 & 90.80 & 72.08 & 91.86 & 90.76 & 40.38 &    \uag{0.81} 78.83 \\
  $\,$ + \textsc{Partial}\textsc{INLP} & 56.27 & 85.43 & 91.39 & 91.58 & 90.68 & 71.60 & 92.51 & 90.70 & 46.01 &    \uag{1.55} 79.57 \\
        $\,$ + \textsc{AM}\textsc{SAL} & 55.51 & 85.28 & 91.35 & 91.36 & 90.62 & 73.41 & 91.67 & 90.61 & 38.50 &    \uag{0.67} 78.70 \\
 $\,$ + \textsc{Kmeans} + \textsc{SAL} & 55.45 & 85.54 & 91.33 & 91.53 & 90.66 & 73.16 & 91.97 & 90.69 & 39.91 &    \uag{0.89} 78.92 \\
    $\,$ + \textsc{Oracle}\textsc{SAL} & 54.50 & 85.44 & 92.09 & 91.78 & 90.70 & 72.68 & 92.05 & 90.79 & 43.19 &    \uag{1.22} 79.25 \\
   $\,$ + \textsc{Partial}\textsc{SAL} & 56.68 & 85.37 & 91.31 & 91.51 & 90.80 & 69.19 & 92.35 & 90.59 & 43.19 &    \uag{0.98} 79.00 \\

\midrule

                             RoBERTa & 58.38 & 87.63 & 92.06 & 92.64 & 91.28 & 71.12 & 94.15 & 90.22 & 52.58 &               81.12 \\
    $\,$ + \textsc{AM} + \textsc{INLP} & 57.91 & 87.58 & 91.66 & 92.57 & 91.31 & 71.48 & 94.15 & 90.10 & 52.11 &    \dab{0.13} 80.99 \\
$\,$ + \textsc{Kmeans} + \textsc{INLP} & 57.37 & 87.51 & 91.47 & 92.83 & 91.26 & 69.92 & 94.15 & 89.96 & 56.34 &    \uag{0.08} 81.20 \\
   $\,$ + \textsc{Oracle}\textsc{INLP} & 56.26 & 87.50 & 91.85 & 92.75 & 91.34 & 72.08 & 94.57 & 90.12 & 52.11 &    \dab{0.16} 80.95 \\
  $\,$ + \textsc{Partial}\textsc{INLP} & 57.53 & 87.70 & 92.26 & 92.68 & 91.31 & 69.80 & 94.19 & 90.15 & 56.34 &    \uag{0.21} 81.33 \\
        $\,$ + \textsc{AM}\textsc{SAL} & 57.47 & 87.63 & 91.50 & 92.81 & 91.31 & 69.19 & 94.30 & 90.01 & 56.34 &    \uag{0.06} 81.17 \\
 $\,$ + \textsc{Kmeans} + \textsc{SAL} & 58.23 & 87.56 & 92.51 & 92.97 & 91.31 & 69.92 & 94.00 & 89.96 & 53.99 &    \uag{0.04} 81.16 \\
    $\,$ + \textsc{Oracle}\textsc{SAL} & 58.47 & 87.85 & 92.22 & 92.84 & 91.27 & 70.76 & 94.50 & 89.94 & 52.11 &    \dab{0.01} 81.11 \\
   $\,$ + \textsc{Partial}\textsc{SAL} & 58.05 & 87.52 & 91.25 & 92.95 & 91.23 & 69.19 & 94.27 & 90.06 & 52.11 &    \dab{0.38} 80.74 \\

\midrule

                               GPT-2 & 32.73 & 82.69 & 84.06 & 87.80 & 89.20 & 65.46 & 92.16 & 84.50 & 40.85 &               73.27 \\
    $\,$ + \textsc{AM} + \textsc{INLP} & 33.78 & 82.66 & 83.91 & 87.77 & 89.15 & 64.86 & 92.13 & 84.47 & 40.85 &    \uag{0.01} 73.29 \\
$\,$ + \textsc{Kmeans} + \textsc{INLP} & 32.67 & 82.68 & 84.36 & 87.83 & 89.16 & 65.46 & 92.13 & 84.51 & 40.85 &    \uag{0.02} 73.29 \\
   $\,$ + \textsc{Oracle}\textsc{INLP} & 33.00 & 82.67 & 84.16 & 87.82 & 89.19 & 65.46 & 92.20 & 84.51 & 40.38 &    \dab{0.01} 73.27 \\
  $\,$ + \textsc{Partial}\textsc{INLP} & 34.51 & 82.56 & 84.22 & 87.88 & 89.17 & 65.34 & 91.78 & 84.10 & 40.85 &    \uag{0.11} 73.38 \\
        $\,$ + \textsc{AM}\textsc{SAL} & 35.07 & 82.77 & 84.86 & 88.21 & 89.13 & 65.94 & 92.28 & 83.93 & 40.38 &    \uag{0.35} 73.62 \\
 $\,$ + \textsc{Kmeans} + \textsc{SAL} & 35.57 & 82.78 & 84.80 & 88.27 & 89.16 & 65.34 & 92.13 & 83.93 & 38.97 &    \uag{0.17} 73.44 \\
    $\,$ + \textsc{Oracle}\textsc{SAL} & 37.23 & 82.77 & 84.64 & 88.39 & 89.14 & 64.50 & 92.05 & 84.02 & 40.85 &    \uag{0.46} 73.73 \\
   $\,$ + \textsc{Partial}\textsc{SAL} & 37.66 & 82.71 & 85.20 & 88.28 & 89.21 & 66.55 & 92.13 & 84.15 & 39.91 &    \uag{0.71} 73.98 \\

\bottomrule
\end{tabular}
}
\caption{\label{tab:glue}GLUE tests for gender-debiased BERT, ALBERT, RoBERTa, and GPT-2 Models.}
\end{table*}

}

\end{document}